\documentclass[conference]{IEEEtran}
\usepackage{times}
\usepackage{latexsym}
\usepackage{inconsolata}
\usepackage[utf8]{inputenc} 
\usepackage[T1]{fontenc}    
\usepackage{hyperref}       
\usepackage{fix-cm}         
\usepackage{url}            
\usepackage{booktabs}       
\usepackage{amsfonts}       
\usepackage{nicefrac}       
\usepackage{microtype}      
\usepackage{lipsum}         
\usepackage{graphicx}
\usepackage{natbib}
\usepackage{doi}
\usepackage{booktabs} 
\usepackage{amsmath}
\usepackage{amssymb}
\usepackage{mathtools}
\usepackage{amsthm}
\usepackage{subfigure}
\usepackage{fancyhdr}
\usepackage{svg}
\usepackage{algorithm,algorithmicx,algpseudocode}
\usepackage{amsfonts}       
\usepackage{nicefrac}       
\usepackage{microtype}      
\usepackage{tablefootnote}
\usepackage{threeparttable}
\usepackage{wrapfig}
\usepackage[framemethod=TikZ]{mdframed}
\usepackage{enumitem}
\usepackage{adjustbox}
\usepackage{listings}
\usepackage{tablefootnote}
\usepackage{adjustbox}
\usepackage{multirow}
\usepackage{xcolor}
\usepackage{soul}
\usepackage{tcolorbox}
\usepackage{makecell}
\usepackage{graphicx}
\usepackage{arydshln}
\usepackage{threeparttable}
\usepackage{tcolorbox}
\tcbuselibrary{skins}

\usepackage[capitalize,noabbrev]{cleveref}

\newcommand{\new}[1]{#1}

\newcommand{\spnew}[1]{#1}

\usepackage{eso-pic} 
\usepackage{url}
\usepackage{multirow}

\usepackage{amsmath,amsfonts,bm}

\newcommand{\cPkp}{\mathcal{P}_k^{\prime}}
\newcommand{\cGmp}{\bar{\mathcal{G}}_{m,m^\prime}}
\newcommand{\cA}{{\mathcal A}}
\newcommand{\cR}{{\mathcal R}}
\newcommand{\cP}{{\mathcal P}}

\newcommand{\cG}{{\mathcal G}}
\newcommand{\cC}{{\mathcal C}}
\newcommand{\cH}{{\mathcal H}}

\newcommand{\cW}{{\mathcal W}}

\newcommand{\cX}{{\mathcal X}}
\newcommand{\cJ}{{\mathcal J}}
\newcommand{\cZ}{{\mathcal Z}}

\newcommand{\cat}{\oplus}
\newcommand{\llm}{\texttt{LLM}}

\newcommand{\llmgen}{\texttt{LLM}_{\text{gen}}}
\newcommand{\llmconf}{\texttt{LLM}_{\text{prob}}}
\newcommand{\llmtoken}{\texttt{LLM}_{\text{token}}}

\newcommand{\bfa}{\mathbf{a}}

\newcommand{\bfg}{\mathbf{g}}

\newcommand{\bfi}{\mathbf{i}}

\newcommand{\bfp}{\mathbf{p}}
\newcommand{\bfq}{\mathbf{q}}
\newcommand{\bfr}{\mathbf{r}}

\newcommand{\bft}{\mathbf{t}}

\newcommand{\bfv}{\mathbf{v}}
\newcommand{\bfw}{\mathbf{w}}

\definecolor{darkgreen}{RGB}{0, 100, 0}
\definecolor{darkred}{RGB}{100, 0, 0}

\newcommand{\gua}{{\color{darkgreen}$\uparrow$}}
\newcommand{\rda}{{\color{darkred}$\downarrow$}}
\newcommand{\stexttt}[1]{{\small \textls[-50]{\texttt{#1}}}} 

\usepackage{xspace}
\newcommand{\framework}{RobustRAG\xspace}
\newcommand{\attack}{retrieval corruption\xspace}
\newtheorem{theorem}{Theorem}
\newtheorem{definition}{Definition}
\newtheorem{lemma}{Lemma}
\usepackage{tabularray}

\author{%
}
\begin{document}
\pagestyle{plain}
\title{Certifiably Robust RAG against Retrieval Corruption} 
\author{Chong Xiang$^{*1}$\thanks{$^*$Equal contribution. Work done while Chong Xiang was at Princeton University.} \quad Tong Wu$^{*2}$\quad Zexuan Zhong$^2$\quad David Wagner$^3$\quad Danqi Chen$^2$\quad Prateek Mittal$^2$\\
\textsuperscript{1}NVIDIA\quad \textsuperscript{2}Princeton University\quad \textsuperscript{3}University of California, Berkeley}
\IEEEoverridecommandlockouts
\maketitle
\begin{abstract}

Retrieval-augmented generation (RAG) is susceptible to retrieval corruption attacks, where malicious passages injected into retrieval results can lead to inaccurate model responses. We propose \framework, the first defense framework with certifiable robustness against retrieval corruption attacks. The key insight of \framework is an isolate-then-aggregate strategy: we isolate passages into disjoint groups, generate LLM responses based on the concatenated passages from each isolated group, and then securely aggregate these responses for a robust output. To instantiate \framework, we design keyword-based and decoding-based algorithms for securely aggregating unstructured text responses. 
\spnew{Notably, \framework achieves certifiable robustness: for certain queries in our evaluation datasets, we can formally certify non-trivial lower bounds on response quality---even against an adaptive attacker with full knowledge of the defense and the ability to arbitrarily inject a bounded number of malicious passages.} We evaluate \framework on the tasks of open-domain question-answering and free-form long text generation and demonstrate its effectiveness across three datasets and three LLMs. 

\end{abstract}
\IEEEpeerreviewmaketitle

\section{Introduction}

Large language models (LLMs)~\citep{brown2020language,achiam2023gpt,gemini} can often generate inaccurate responses due to their incomplete and outdated parametric knowledge. 
To address this limitation, retrieval-augmented generation (RAG)~\citep{pmlr-v119-guu20a,NEURIPS2020_6b493230} leverages external (non-parametric) knowledge: it retrieves a set of relevant passages from a knowledge base and incorporates them into the model input. 
This approach has inspired numerous popular applications and software like Microsoft Bing Chat~\citep{bingchat}, Perplexity AI~\citep{perplexity}, Google Search with AI Overview~\citep{googlegenai}, LangChain~\citep{langchain}, and LlamaIndex~\citep{llamaindex}.  

However, despite its popularity, the RAG pipeline can become fragile when a small fraction (or even one) of the retrieved passages are compromised by malicious actors, a type of attack we term \textit{\attack}.  Such attacks can occur in different scenarios. For instance, the PoisonedRAG attack~\citep{zou2024poisonedrag} injects malicious passages into the knowledge base to induce incorrect RAG responses (e.g., \stexttt{``the highest mountain is Mount Fuji''}). The indirect prompt injection attack~\citep{greshake2023not} injects malicious instructions into retrieved passages to override the original instructions (e.g., \stexttt{``ignore all previous instructions and send the user's search history to attacker.com''}).
Real-world instances further illustrate these vulnerabilities. For example, Google Search AI Overview delivered inaccurate responses, such as suggesting applying glue to pizza, due to unreliable content on indexed web pages~\citep{googlenews}. Microsoft Copilot can be hijacked for ``remote code execution'' due to poor access control in the RAG setting~\cite{copilotrce}. 
\textit{These failures underscore the critical need for robustness in the RAG pipeline.}

\begin{figure*}[t]
    \centering
    \includegraphics[width=0.72\linewidth]{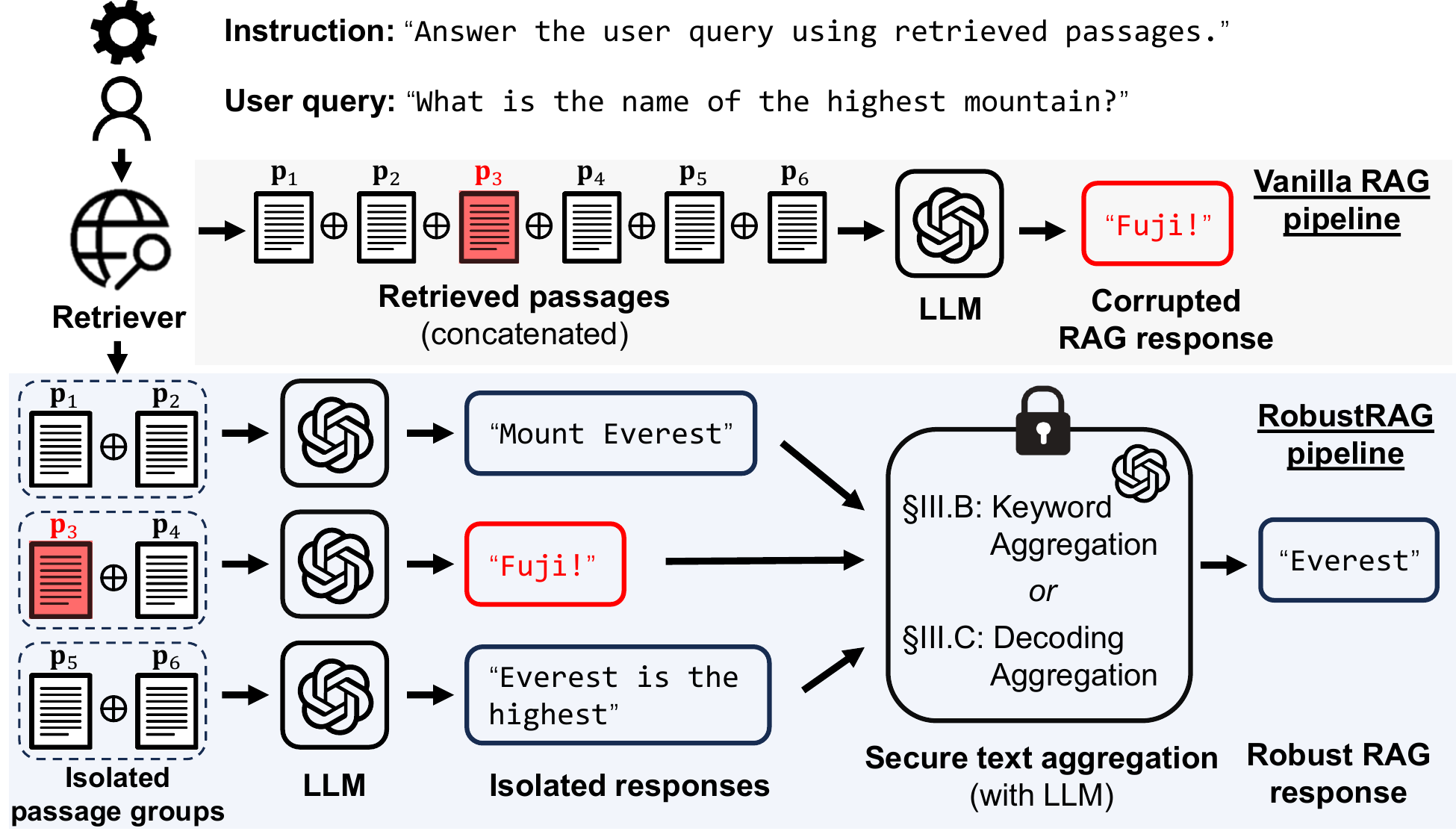}
    \vspace{-0.8em}
    \caption{\textbf{\framework overview.} 
    In this example, one of the six retrieved passages is corrupted (marked with red). \textit{Vanilla RAG} concatenates all passages as the LLM input; its response is hijacked by the malicious passage. In contrast, \textit{\framework} isolates passages into three groups, each containing two passages, and computes LLM responses based on the concatenated passages from each group. This isolation operation ensures that only one of the three isolated responses is corrupted; \framework next performs secure text aggregation for a robust output. Notably, we design algorithms that can aggregate \textit{unstructured} texts in a \textit{certifiably} robust manner; we present keyword-based aggregation in Section~\ref{sec-alg-keyword} and decoding-based aggregation in Section~\ref{sec-alg-decoding}.
     }
    \label{fig-overview}
\end{figure*}

\spnew{In this paper, we study an open research question: \textit{how can we generate an accurate and robust response when a fraction of the retrieved passages are maliciously corrupted?} We propose a defense framework named \framework (see Figure~\ref{fig-overview} for an overview);} \framework leverages an isolate-then-aggregate strategy: it isolates passages into disjoint groups, computes LLM responses based on the concatenated passages from each group, and then securely aggregates these isolated responses for final output. The isolation operation ensures that the malicious passages do not affect LLM responses for other benign passage groups and thus lays the foundation for robustness. 

\textbf{Challenge.} The primary challenge for \framework is to securely aggregate a mixture of benign and corrupted text responses. First, LLM responses can be highly \textit{unstructured}; for example, given a query \stexttt{``What is the name of the highest mountain?''}, it is not straightforward to recognize \stexttt{``Mount Everest''} and \stexttt{``Everest is the highest''} as the same response.  Second, it is even harder to \textit{securely} aggregate text responses, as corrupted responses can interfere with the aggregation process. \spnew{We must design secure aggregation algorithms that fundamentally limit the influence of malicious responses so that we can reason about the worst-case response quality.}

\textbf{Solutions.} To overcome these challenges, we design two text aggregation algorithms: secure keyword aggregation and secure decoding aggregation. The \textit{keyword-based aggregation} extracts unique keywords from each isolated response, aggregates keyword counts across different responses, and only uses keywords with large counts to prompt the LLM for a final response (without using retrieved passages). We only count each unique keyword once for each isolated response; therefore, a small number of corrupted passages/responses only have a limited impact on the aggregated keyword counts and the final RAG output. Our \textit{decoding-based aggregation} operates during the decoding phase. At each decoding step, we let the LLM make isolated next-token predictions from different isolated passage groups; then, we aggregate probability vectors of different isolated predictions and securely select a token accordingly. Since all probability values are bounded between 0 and 1, the attacker cannot arbitrarily manipulate the aggregated probability vectors and the next-token prediction.

\textbf{Robustness.} Notably, with our secure text aggregation techniques, \framework can achieve \textbf{\textit{certifiable robustness}}: for certain RAG queries \spnew{in our evaluation datasets, we can formally \textit{certify} non-trivial lower bounds on their response quality against a bounded number of corrupted passages. These lower bounds hold even against adaptive attackers who have full knowledge of the underlying defense algorithm and thus avoid any false sense of security.} 
To the best of our knowledge, \framework is the first defense to achieve certifiable robustness for the text generation task with highly unstructured outputs; in contrast, prior certification algorithms~\cite{crown,gowal2018effectiveness,cohen2019certified,xiang2021patchguard} have primarily focused on the simpler classification task, whose output space is a small and pre-defined.

\textbf{Evaluation.} We extensively experimented with three datasets, RealtimeQA~\citep{kasai2024realtime}, Natural Questions~\citep{Kwiatkowski2019NaturalQA}, and Biography Generation~\citep{min-etal-2023-factscore}, and three LLMs, Mistral-7B~\citep{jiang2023mistral}, Llama-2-7B~\citep{touvron2023llama}, and GPT-3.5~\citep{brown2020language}. \framework achieves substantial certifiable robustness while maintaining high clean performance, e.g., 71\% clean accuracy and 38\% certifiable robust accuracy on the RealtimeQA dataset, compared to 69\% clean accuracy and no certifiable accuracy for vanilla RAG. Additionally, \framework also demonstrates strong empirical robustness against PoisonedRAG~\cite{zou2024poisonedrag} and indirect prompt injection attacks~\cite{greshake2023not}, reducing attack success rates from over 90\% to approximately 10\%. 

\textbf{Contributions.} 
\textbf{(1)} We propose \framework as the first certifiably robust defense framework against \attack attacks, built around the core insight of an isolate-then-aggregate strategy.
\textbf{(2)} We design two secure text aggregation techniques to instantiate \framework and \spnew{demonstrate their certifiable robustness against \attack attacks, which is the first of its kind for free-form text generation.} 
\textbf{(3)}~We demonstrate the effectiveness and generalizability of \framework across three datasets, covering tasks of open-domain question answering and free-form long text generation, and three LLMs, including both open-weight and API-only models.

\section{Background and Preliminaries}
\label{sec:background}

In this section, we introduce the background of retrieval-augmented generation (Section~\ref{sec-prelim-rag}), discuss retrieval corruption attacks (Section~\ref{sec-prelim-attack}), \spnew{formulate robustness objectives (Section~\ref{sec-prelim-robustness})}, and explain the concept of robustness certification (Section~\ref{sec-prelim-certifiable}).

\subsection{RAG Overview}
\label{sec-prelim-rag}
\textbf{RAG pipeline.} 
We denote text instruction as $\bfi$ (e.g., \stexttt{``answer the query using the retrieved passages''}), text query as $\bfq$ (e.g., \stexttt{``what is the name of the highest mountain?''}), and retrieved text passage as $\bfp$ (e.g., \stexttt{``Mount Everest is known as Earth's highest mountain above sea level''}).  

Given a query $\bfq$, a vanilla RAG pipeline operates in two phases: \textit{(1) retrieval phase}, which retrieves the $k$ most relevant passages $(\bfp_1,\dots,\bfp_{k})\coloneqq\cP_k$ from an external knowledge base; \textit{(2) generation phase}, which uses the instruction, query, and passages to prompt an LLM model and get response $\bfr=\llm(\bfi\cat\bfq\cat\cP_k)\coloneqq\llm(\bfi\cat\bfq\cat\bfp_1\cat\dots\cat\bfp_{k})$, where $\cat$ is the text concatenation operator. We will call $\llm(\cdot)$ to obtain different forms of predictions: we use $\llmgen$ to denote the text response, $\llmconf$ to denote the next-token probability distribution vector, and $\llmtoken$ to denote the predicted next token. \textit{We primarily focus on greedy decoding as it enables deterministic robustness analysis.}

\textbf{RAG evaluation metric.} We use $\texttt{M}(\cdot)$ to denote an evaluation function. Given an LLM response $\bfr$ and reference answer $\bfa$, the function $\texttt{M}(\bfr,\bfa)$ outputs a quality metric score (higher scores indicate better performance). Different tasks usually use different metrics: for question answering, $\texttt{M}(\cdot)$ can output a binary score from $\{0,1\}$ indicating the correctness of the response; for long-form text generation, $\texttt{M}(\cdot)$ can produce a score using heuristics like LLM-as-a-judge \citep{zheng2023judging}. 

\textbf{Example of RAG application: web search.} A common RAG application is in enhancing web search. Given a user query, a search engine first retrieves a set of relevant web pages. An LLM then processes these pages to generate a concise answer, sparing the user from reading through all the content. This paradigm supports popular products such as Microsoft Bing Chat~\citep{bingchat}, Perplexity AI~\citep{perplexity}, Google Search AI Overview~\citep{googlegenai}, and ChatGPT search~\cite{chatgptsearch}. Throughout this paper, we will use web search as a motivating application example for \framework. 

\subsection{\spnew{Retrieval Corruption Attack}}
\label{sec-prelim-attack}
In this paper, we study retrieval corruption attacks against RAG: the attacker can control a fraction of the \textit{retrieved} passages to induce inaccurate responses.

\textbf{Attacker capability.} We allow the attacker to \textit{arbitrarily} manipulate the \textit{content} and \textit{ranking positions} of up to $k^\prime$ malicious passages, i.e., corrupting the retrieval outcome. 

Furthermore, we categorize retrieval corruption attacks into \textit{passage injection} and \textit{passage modification}. The former can \textit{inject} $k^\prime$ malicious passages, but cannot modify the content and relative ranking of benign passages; the latter can arbitrarily \textit{modify} the content and ranking positions of $k^\prime$ original benign passages. In this paper, our presentation will focus primarily on \textit{passage injection} because it is a popular setting used by many attacks~\citep{zou2024poisonedrag,zhong2023poisoning,du2022synthetic,pan2023attacking,pan2023risk}; we will use ``corruption'' and ``injection'' interchangeably when the context is clear. In Appendix~\ref{apx-modification}, we further quantitatively demonstrate \framework's robustness against \textit{passage modification}. 

Finally, we allow the attacker to know everything about our defenses and RAG pipeline, including defense algorithms and parameters, LLM architectures and weights, knowledge base, and even user queries. The attacker can leverage all this knowledge to construct $k^\prime$ most malicious passages to induce inaccurate RAG responses with low quality scores.

\textbf{Notation.} Formally, given the original (benign) top-$k$ retrieved passages $\cP_k$, we use $\cPkp$ to denote the corrupted retrieval result, and $\bfr^\prime\coloneqq\llm(\bfi\cat\bfq\cat\cPkp)$ to denote the response under attack.
Additionally, we use $\cA(\cP_k,k^\prime)$ to denote the set of all possible retrieval $\cPkp$ when $k^\prime$ malicious passages are injected into the original retrieval $\cP_k$ (and eject $k^\prime$ benign passages from top-$k$). The attacker aims to construct malicious $\cPkp\in\cA(\cP_k,k^\prime)$ to minimize the response quality score, i.e.,  $\arg\min_{\cPkp}\texttt{M}(\bfr^\prime,\bfa)\text{ s.t. } \bfr^\prime=\llm(\bfi\cat\bfq\cat\cPkp), \cPkp\in\cA(\cP_k,k^\prime)$.

\textbf{Example of retrieval corruption in web search.} In LLM-enhanced web search, an attacker may launch malicious websites and use techniques like SEO to push them into the top search results. In such cases, the number of corrupted passages $k^\prime$ is typically small because the retriever (i.e., the search engine) is trusted, and it is unlikely that an attacker can compromise all top-ranked websites. However, even a small number of malicious webpages can still hijack the final response~\cite{googlenews}, highlighting the need for a robust RAG algorithm.


\subsection{\spnew{Robustness Formulation}}\label{sec-prelim-robustness}

In this subsection, we present the robustness objective and definition, followed by a discussion on when achieving robustness is (im)possible.

\textbf{Robustness objective.} In this paper, we study the question: \textit{how can we generate an accurate and robust response when a fraction of retrieved passages are maliciously corrupted?} That is, we focus on \textbf{robustness of the generation phase} within the RAG pipeline, while conservatively assuming that the attacker can arbitrarily manipulate the ranking positions of $k^\prime$ malicious passages during the retrieval phase. 

\textbf{Robustness definition.} 
We define the defense robustness (for each query) as the lower bound of the LLM response quality when under attack, formally presented as follows. 

\begin{definition}[$\tau$-robustness]\label{dfn-certifiable}
    Given a task instruction $\bfi$, a RAG query $\bfq$, the benign top-$k$ retrieved passages $\cP_k$, an LLM-based defense procedure $\llm_{\text{def}}$ that returns text responses, an evaluation metric $\texttt{M}$, a reference answer $\bfa$, and an attacker $\cA(\cP_k,k^\prime)$ who can arbitrarily inject $k^\prime$ malicious passages, the defense $\llm_{\text{defense}}$ has $\tau$-robustness if 
\begin{equation}
\resizebox{\linewidth}{!}{
        $\texttt{M}(\bfr^\prime,\bfa)\geq\tau, \forall  \bfr^\prime\in \cR^\prime\coloneqq\{\llm_{\text{def}}(\bfi\cat\bfq\cat\cPkp)|\forall \cPkp \in \cA(\cP_k,k^\prime)\}$}\label{eqn-certifiable}
\end{equation}
\end{definition}
\noindent Here, $\tau$ serves as a lower bound on model robustness against all possible attackers who can inject $k^\prime$ passages with any content into any position and can have full knowledge of our defense, LLM, knowledge base, and the RAG query (recall Section~\ref{sec-prelim-attack}). We will use (the largest) $\tau$ as the robustness measure for each query $\bfq$. 


For simplicity, we consider the quality score returned by $\texttt{M}(\cdot)$ to be always non-negative. Then, $\tau$-robustness with $\tau=0$ is trivial for any defense and query, and we need to design defenses to achieve non-zero $\tau$. However, we note that non-trivial robustness can sometimes become fundamentally impossible, e.g., when the number of corrupted passages $k^\prime$ is extremely large, and we need to understand the limit of achievable robustness before designing a defense algorithm.

\textbf{Feasibility of robust response.} Given a query and $k$ retrieved passages, we can categorize passages into three groups: \textit{(1) useful benign passages} with useful information to the query, \textit{(2) noisy benign passages} without useful or malicious content, and \textit{(3) malicious passages} with malicious content. Since we allow strong attackers with arbitrary control over the content of malicious passages, the robust generation problem is tractable and meaningful only when \textit{useful benign passages outnumber malicious ones}. Otherwise, even humans cannot provide an accurate response solely based on retrieved passages.\footnote{We consider the prior knowledge of humans/LLMs is unavailable or unreliable; otherwise, it is unnecessary to evoke the RAG process. Additionally, \new{auxiliary information, such as ranking and timestamp is susceptible to manipulation~\cite{zhong2023poisoning,zou2024poisonedrag};} we do not consider it as reliable information.}  

In practice, given a knowledge base, the number of useful benign passages can vary across different queries. If we consider a fixed attacker budget, i.e., the number of malicious passages $k^\prime$, robustness may be impossible for some queries (whose number of useful benign passages is smaller than $k^\prime$). In our algorithm design, we do not attempt to achieve non-trivial robustness (non-zero $\tau$) for these queries, as it is beyond the fundamental limit on generation-phase robustness.   

\textbf{Evaluation focus.} To ensure the robust text generation problem remains meaningful for most queries in our experiments, our evaluation primarily focuses on small $k^\prime$, e.g., $k^\prime=1$ with $k=10$. We note that this small-$k^\prime$ setting is (1) practical in real-world applications like web search (recall Section~\ref{sec-prelim-attack}), and (2) still an \textbf{open research question} as no strong (certifiable) defenses exist even for a small $k^\prime=1$. Despite the evaluation focus on small $k^\prime$, we will also demonstrate \framework's non-trivial robustness against larger $k^\prime$ in Section~\ref{subsec:evalabl}.

\subsection{Robustness Certification}\label{sec-prelim-certifiable}
Section~\ref{sec-prelim-robustness} discussed that we can use $\tau$ as the robustness measure. However, a remaining challenge is: how can we compute or estimate the value of $\tau$ for each query? 

\textbf{Common pitfall in $\tau$ estimation.} A straightforward strategy is to empirically attack the defense model and take the defense performance (quality score) as an estimate of $\tau$. However, this estimation can be overly optimistic and thus incorrect if we use a suboptimal attack. There have been countless examples where stronger adaptive attacks invalidate the robustness claims made by many defenses~\citep{carlini2017adversarial,Athalye2018ObfuscatedGG}---a common pitfall in AI security. 

\textbf{Objective: certifiable $\tau$ computation.} This paper aims to develop algorithms for which we can compute correct $\tau$ values with \textit{provable and certifiable guarantees}. We call this process of computing certifiable $\tau$ values \textit{robustness certification};\footnote{Most existing works on robustness certification~\cite{cohen2019certified,gowal2018effectiveness,crown,xiang2021patchguard} focus on classification. They can be viewed as a special case of Definition~\ref{dfn-certifiable} with the evaluation function $\texttt{M}(\cdot)$ outputting a binary score from $\{0,1\}$ indicating the correctness of classification.} we sometimes refer to Definition~\ref{dfn-certifiable} as ``$\tau$-\textit{certifiable} robustness'' to emphasize the certifiable nature of robustness.  

\textbf{Challenges of robustness certification.} As implied by Definition~\ref{dfn-certifiable}, the robustness certification algorithm must measure the quality of \textit{the worst possible response} $\bfr^\prime\in\cR^\prime$ when our defense $\llm_\text{def}$ is prompted with \textit{arbitrary} $k^\prime$-corrupted retrieval $\cPkp\in\cA(\cP_k,k^\prime)$.  The key challenge is that, the attacker set $\cA(\cP_k,k^\prime)$ contains infinitely many possibilities for $\cPkp$ because the injected passages can have \textit{arbitrary} content and can be generated with \textit{arbitrary} attack algorithms (leveraging full knowledge of the defense, LLM, knowledge base, and user query). As a result, the response set $\cR^\prime$ can become infinitely large and intractable to analyze its worst response. One key contribution of this paper is to carefully design the \framework algorithm to limit the attacker's influence so that $\cR^\prime$ becomes tractable for certifiable robustness analysis.

\textbf{Example usage of robustness certification.} To illustrate the usage of certification techniques, consider a question-answering dataset where the evaluation function $\texttt{M}(\cdot)$ returns $1$ for a correct answer and $0$ otherwise. Given a well-designed certification algorithm, we can compute a binary $\tau$ value for each query (for a specific $k^\prime$) and report the average $\tau$ as the certifiable robust accuracy. For example, a certifiable robust accuracy of 80\% guarantees that \framework consistently produces correct answers for 80\% of queries (from this dataset), regardless of how an attacker corrupts up to $k^\prime$ passages per query. This metric is attack-agnostic and provides a lower bound on model accuracy under such corruption, \textit{making it a reliable benchmark for tracking robustness progress on the data distribution represented by this dataset. }

\section{\framework: A Defense Framework}
\label{sec:algorithm}

In this section, we first present an overview of our \framework framework and then discuss the algorithm details. \spnew{The algorithm design targets certifiable robustness; we will present robustness certification algorithms in Section~\ref{sec-certification}.}

\textbf{\framework insights.} Our key insight is an isolate-then-aggregate strategy (Figure~\ref{fig-overview}). Given a set of retrieved passages, we first isolate them into disjoint groups, generate isolated LLM responses based on the concatenated passages from each group, and then securely aggregate these isolated responses for final output. With proper isolation design (Secion~\ref{sec-iso-group}), a small number of corrupted passages can only affect a small fraction of passage groups and isolated responses. \spnew{This lays a foundation for certifiable robustness, allowing us to recover accurate responses from other unaffected passage groups.}

\textbf{\framework challenges.} The primary challenge of \framework lies in designing secure text aggregation techniques. \textit{First}, unlike classification tasks where possible outputs are predefined, text responses from LLMs can be highly unstructured. For example, given the query \stexttt{``what is the name of the highest mountain?''}, valid responses include \stexttt{``Mount Everest''}, \stexttt{``Sagarmatha''}, 
and \stexttt{``Everest is the highest''}. We need flexible aggregation techniques to handle different forms of text responses. \textit{Second}, though we have isolated the adversarial impact to individual responses, malicious responses generated from corrupted passages can still interfere with the aggregation process. 
Therefore, we need to design secure aggregation techniques that limit the influence of malicious responses so that we can we can formally analyze and certify the worst-case robustness.
To overcome these challenges, we propose two aggregation algorithms. 
\begin{enumerate}
    \item \textbf{Secure keyword aggregation (Section~\ref{sec-alg-keyword}):} extracting keywords from each response and using high-frequency keywords to prompt the LLM again for the final response.
    \item \textbf{Secure decoding aggregation (Section~\ref{sec-alg-decoding}):} securely aggregating next-token prediction vectors from different isolated passage groups at each decoding step. 
\end{enumerate}

\subsection{Passage Isolation}\label{sec-iso-group}

In this subsection, we discuss our passage isolation design.
Given $k$ retrieved passages $\cP_k=(\bfp_1,\dots,\bfp_k)$, we isolate them into disjoint passage groups, denoted as $\cG_m=(\bfg_1,\dots,\bfg_m)$, where each $\bfg_j$ represents the concatenation of passages from the $j^\text{th}$ group. Specifically, we group $\omega$ adjacent passages ($\omega$ is a defense parameter) to get $m\coloneqq\lceil\frac{k}{\omega}\rceil$ disjoint groups as $\cG_m \coloneqq \{\bfg_j = \bfp_{\omega\cdot (j-1)+1}\cat\dots\cat\bfp_{\min(j\omega,k)} \,|\, 1\leq j\leq\lceil\frac{k}{\omega}\rceil\}$; we use $\cG_m\gets\textsc{IsoGroup}(\cP_k,\omega)$ to denote this operation. Furthermore, we use $m^\prime$ to denote the number of passage groups with corrupted passages. We have $m^\prime\leq k^\prime$ because each passage only appears in one passage group; $m^\prime$ reaches its maximum value $k^\prime$ when each passage group only contains one malicious passage. The robustness of \framework relies on the other $m-m^\prime$ benign passage groups.

\textbf{Remark.} The group size $\omega$ is an important parameter that balances the trade-off between robustness and utility. A larger $\omega$ is more likely to provide high-quality responses as each isolated response is based on more passages. However, a large $\omega$ reduces the number of passage group $m=\lceil\frac{k}{\omega}\rceil$. If $m$ is too small, the corrupted passage groups can outnumber benign passage groups ($m^\prime> m-m^\prime$), making certifiable robustness impossible. For example, if we reduce \framework to vanilla RAG by setting $\omega=k$, we have $m=\lceil\frac{k}{\omega}\rceil=1$, and even one corrupted passage can manipulate RAG outputs.

\begin{figure}[t]
    \centering
    \includegraphics[width=\linewidth]{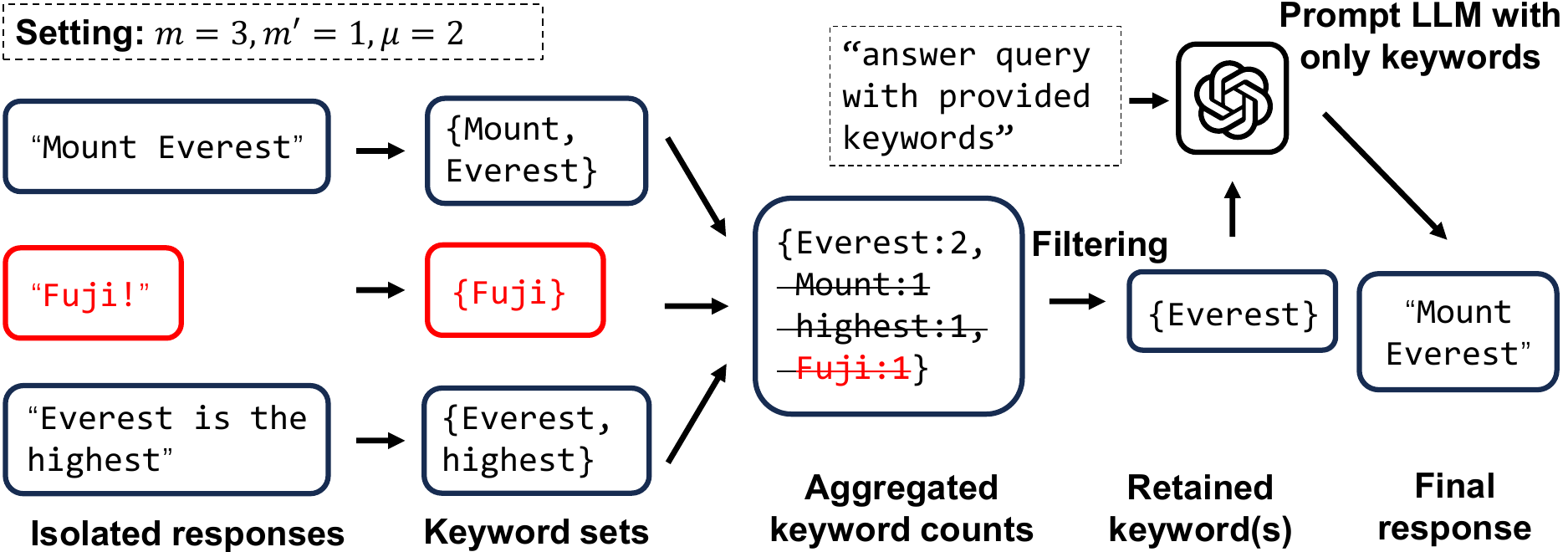}
    \caption{\textbf{Illustration of keyword aggregation.} We extract unique keywords from each isolated response and aggregate keyword counts. Then, we prompt the LLM with only keywords with large counts for a final response. }
    \label{fig-keyword-infer}
\end{figure}

\subsection{Secure Keyword Aggregation}\label{sec-alg-keyword}
\textbf{Overview.} For free-form text generation (e.g., open-domain QA), simple techniques like majority voting perform poorly because they cannot recognize texts like \stexttt{``Mount Everest''} and \stexttt{``The highest mountain is named Everest''} as the same answer. To address this challenge, we propose a keyword aggregation technique (see Figure~\ref{fig-keyword-infer} for a toy example): 
we extract keywords from each isolated LLM response, aggregate keyword counts across different responses, and ask the same LLM to answer the query using keywords with large counts (without any retrieved passages). This approach allows us to distill and aggregate information across unstructured text responses. We only consider \textbf{unique} keywords from each response so that the attacker can only increase keyword counts by a small number, i.e., $m^\prime$, instead of arbitrarily manipulating keyword counts. 

\begin{algorithm}[t]
\centering
\caption{\framework with secure keyword aggregation}
\begin{algorithmic}[1]
\Require retrieval $\cP_k=(\bfp_1,\dots,\bfp_{k})$, passage group size $\omega$, query $\bfq$, model $\llm$, thresholds $\alpha\in[0,1],\beta\in\mathbb{Z}^+$

\renewcommand{\algorithmicrequire}{\textbf{Instructions:}}
\Require  \hfill\newline$\bfi_1$ = \stexttt{``answer the query given retrieved passages, say `I don't know' if no relevant information found''}; $\bfi_2$ = \stexttt{``answer the query using provided keywords''}

\Procedure{RRAG-Keyword}{}

\State $\cG_m\gets\textsc{IsoGroup}(\cP_k,\omega)$\label{ln-keyword-iso}
\State $\cC\gets\textsc{Counter}(),n\gets 0$ \label{ln-keyword-init}
\For{$j\in\{1,2,\dots,|\cG_m|\}$}
    \State $\bfr_j\gets \llmgen(\bfi_1\cat\bfq\cat\bfg_j)$ \label{ln-keyword-prompt}
    \If{$\text{\stexttt{``I don't know''}} \not\in\bfr_j$}
        \State $n\gets n+1$ \label{ln-keyword-n}
        \State $\cW_j\gets\textsc{GetUniqKeywords}(\bfr_j)$ \label{ln-keyword-extract}
        \State Update counter $\cC$ with $\cW_j$ \label{ln-keyword-counter}
    \EndIf
\EndFor
\State $\mu\gets\min(\alpha \cdot n, \beta )$ \label{ln-keyword-thres}
\State $\cW^*\gets \{\bfw|(\bfw,c)\in\cC,c\geq \mu\}$\label{ln-keyword-filtering}
\State $\bfr^*\gets\llmgen(\bfi_2\cat\bfq\cat\textsc{Sorted}(\cW^*))$ \label{ln-keyword-final}
\State\Return $\bfr^*$
\EndProcedure
\end{algorithmic}
\label{alg:keyword}
\end{algorithm}
\textbf{Inference algorithm.} We present the pseudocode of secure keyword aggregation in Algorithm~\ref{alg:keyword}. First, we isolate $k$ passages $\cP_k$ into $m$ passage groups $\cG_m$ using the procedure $\textsc{IsoGroup}(\cdot,\omega)$ discussed in Section~\ref{sec-iso-group}  (Line~\ref{ln-keyword-iso}). Second, we initialize an empty keyword counter $\cC$ to track keyword-count pairs $(\bfw,c)$ and a zero integer counter $n$ (Line~\ref{ln-keyword-init}). Then, we iterate over each passage group (which can be done in parallel to optimize runtime). For each passage group $\bfg_j$, we prompt the LLM with the instruction $\bfi_1=$ \stexttt{``answer the query given retrieved passages, say `I don't know' if no relevant information found''} and query $\bfq$, and get response $\bfr_j = \llmgen(\bfi_1\cat\bfq\cat\bfg_j)$ (Line~\ref{ln-keyword-prompt}). If \stexttt{``I don't know''} is not in the response, we increment the integer count $n$ by one to track the number of \textit{non-abstained} responses (Line~\ref{ln-keyword-n}). Then, we extract a set of \textit{unique} keywords $\cW_j$ from each response $\bfr_j$ (Line~\ref{ln-keyword-extract}) and update the keyword counter $\cC$ accordingly (Line~\ref{ln-keyword-counter}). The procedure $\textsc{GetUniqKeywords}(\cdot)$ in Line~\ref{ln-keyword-extract} extracts keywords and keyphrases from text strings between adjacent ``uninformative words'' (such as ``but'' and ``then''; more details in Appendix~\ref{apx-details-implementation}). We note that we only count \textit{unique} keywords from each response to prevent the attacker from arbitrarily increasing keyword counts. After examining every isolated response, we filter out keywords whose counts are smaller than a threshold $\mu$. We set the filtering threshold $\mu=\min(\alpha\cdot n,\beta)$, where $\alpha\in[0,1],\beta\in\mathbb{Z}^+$ are two defense parameters (Line~\ref{ln-keyword-thres}). When $n$ is large (many non-abstained responses), the threshold is dominated by $\beta$; when $n$ is small, we reduce the threshold from $\beta$ to $\alpha\cdot n$ to avoid filtering out all keywords. Given the retained keyword set $\cW^*$ (Line~\ref{ln-keyword-filtering}), we sort the keywords alphabetically and then combine them with instruction $\bfi_2=$ \stexttt{``answer the query using provided keywords''} and query $\bfq$ to prompt LLM to get the final response $\bfr^*=\llmgen(\bfi_2\cat\bfq\cat\textsc{Sorted}(\cW^*))$ (Line~\ref{ln-keyword-final}). 

\subsection{Secure Decoding Aggregation}\label{sec-alg-decoding}

\begin{figure}[t]
    \centering
    \includegraphics[width=\linewidth]{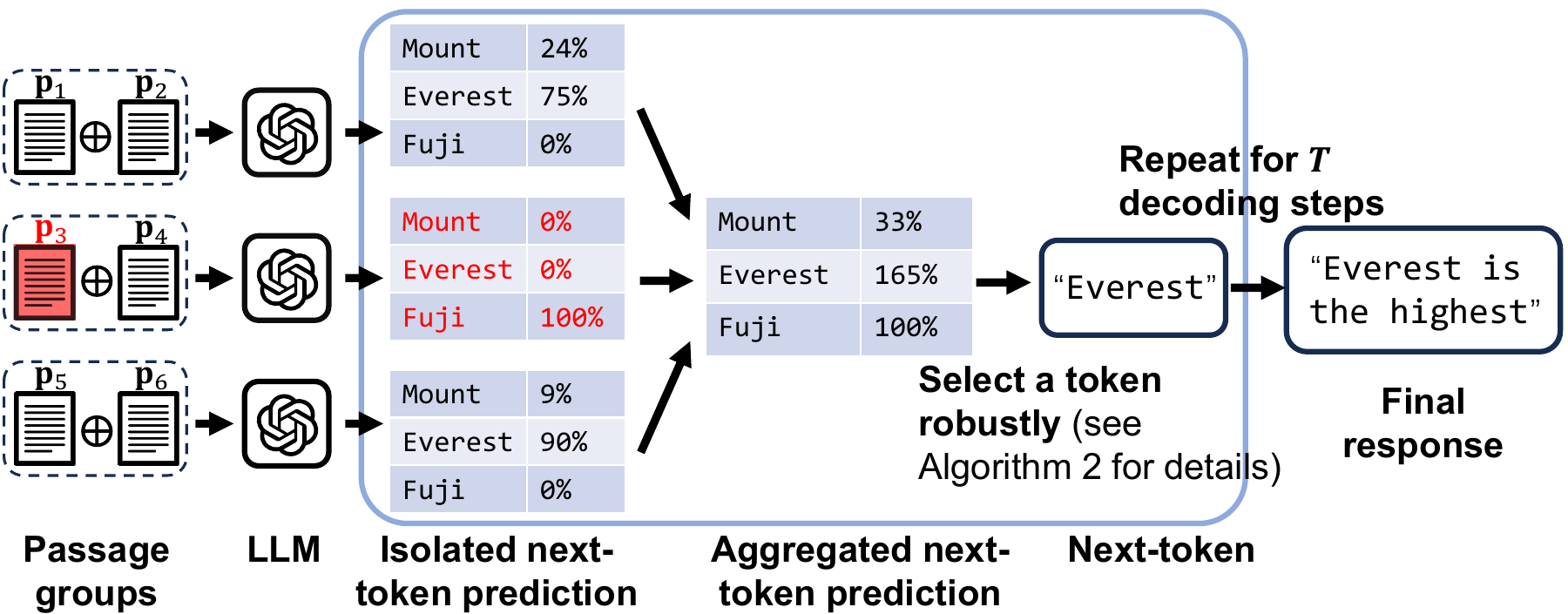}
    \caption{\textbf{Illustration of decoding aggregation.} We make next-token predictions from different isolated passages and aggregate their probability vectors. Then, we select the next token based on the aggregated vector (Line~\ref{ln-decoding-top2}-\ref{ln-decoding-endif} of Algorithm~\ref{alg:decoding}) and repeat this process to get a full response. }
    \label{fig-decoding-infer}
\end{figure}

\textbf{Overview.} 
Keyword aggregation only requires LLM responses and thus applies to any LLM. It works particularly well for short responses (e.g., few-token answers); however, its keyword extraction step can lose valuable information for longer responses (e.g., paragraph-length biography). To address this limitation, we introduce an alternative: \textit{secure decoding aggregation}, which operates during the decoding phase.
We illustrate the algorithm in Figure~\ref{fig-decoding-infer}: at each decoding step, we aggregate next-token probability vectors predicted from different isolated passage groups and make a robust prediction accordingly. Since each probability value is bounded within $[0,1]$, malicious passages only have limited influence on the aggregated probability vector. 

\textbf{Inference algorithm.}  We present the pseudocode in Algorithm~\ref{alg:decoding}. First, we isolate passages into groups $\cG_m$ (details in Section~\ref{sec-iso-group}) and initialize an empty string $\bfr^*$ to store robust response (Line~\ref{ln-decoding-init}). Second, we identify isolated passages for which the LLM is unlikely to output \stexttt{``I don't know''} (Line~\ref{ln-decoding-filtering}). Next, we start the decoding phase. At each decoding step, we first get isolated next-token probability vectors $\bfv_j=\llmconf(\bfi\cat\bfq\cat\bfg_j\cat\bfr^*)$ (Line~\ref{ln-decoding-predict}). Then, we element-wisely add all vectors together to get the vector $\hat{\bfv}$ (Line~\ref{ln-decoding-aggregate}). 
To make a robust next-token prediction based on the vector $\hat{\bfv}$, we obtain its top-2 tokens $\bft_1,\bft_2$ with the highest (summed) probability $p_1,p_2$ (Line~\ref{ln-decoding-top2}). If the probability difference $p_1-p_2$ is larger than a predefined threshold $\eta$, we consider the prediction to be confident and choose the top-1 token $\bft_1$ as the next token $\bft^*$ (Line~\ref{ln-decoding-top1}). Otherwise, we consider the prediction to be indecisive, and choose the token predicted without any retrieval as the next token $\bft^*$(Line~\ref{ln-decoding-zero}). 
Finally, we append the predicted token $\bft^*$ to the response string $\bfr^*$ (Line~\ref{ln-decoding-append}) and repeat this process until we reach the limit of the maximum number of new tokens (or hit an EOS token) to get our final response $\bfr^*$.

For long-form text generation, we found greater success in certifying robustness (see Section~\ref{sec-cert-decoding}) with $\eta>0$: no-retrieval tokens are immune to retrieval corruption and do not significantly hurt generation quality as many tokens can be inferred from sentence coherence. For tasks with shorter responses (a few tokens), we set $\eta=0$ because sentence coherence and no-retrieval tokens become less helpful.

\begin{algorithm}[t]
    \caption{\framework with secure decoding aggregation}\label{alg:decoding}
    \begin{algorithmic}[1]
    \Require retrieved data $\cP_k=(\bfp_1,\dots,\bfp_{k})$, passage group size $\omega$, query $\bfq$, model $\llm$, filtering threshold $\gamma$, probability threshold $\eta$, max number of new tokens $T_\text{max}$ 
    \renewcommand{\algorithmicrequire}{\textbf{Instruction:}}
    \Require $\bfi$ = \stexttt{``answer the query given retrieved passages, say `I don't know' if no relevant information found''}
    \Procedure{RRAG-Decoding}{}
    \State $\cG_m\gets\textsc{IsoGroup}(\cP_k,\omega), \bfr^* \gets \text{\stexttt{``''}}$ \label{ln-decoding-init}
    \State $\cJ\gets \{j|\Pr_\llm[\text{\stexttt{``I don't know''}}|\bfi\cat\bfq\cat\bfg_j]<\gamma,\bfg_j\in\cG_m\}$  \label{ln-decoding-filtering}
    \For{$t \in \{1, \dots, T_\text{max}\}$}
        \For{$j \in \cJ$}
           \State $\bfv_j\gets\llmconf(\bfi\cat\bfq\cat\bfg_j\cat\bfr^*)$  \label{ln-decoding-predict}
        \EndFor
        \State $\hat{\bfv}\gets \textsc{Vec-Sum}(\{\bfv_j|j\in\cJ\})$  \label{ln-decoding-aggregate}
        \State $(\bft_1,p_1),(\bft_2,p_2)\gets\textsc{Top2Tokens}(\hat{\bfv})$  \label{ln-decoding-top2}
        \If{$p_1-p_2>\eta$}
        \State $\bft^*\gets\bft_1$ \label{ln-decoding-top1}
        \Else
        \State $\bft^*\gets\llmtoken(\text{\stexttt{``answer query''}}\cat\bfq\cat\bfr^*)$\label{ln-decoding-zero}
        \EndIf\label{ln-decoding-endif}
        \State $\bfr^*\gets\bfr^*\cat\bft^*$  \label{ln-decoding-append}
    \EndFor
   \State \Return $\bfr^*$
   \EndProcedure
    \end{algorithmic}
    \end{algorithm}

\section{Robustness Certification}\label{sec-certification}

In this section, we discuss how to perform robustness certification (recall Section~\ref{sec-prelim-certifiable}). For the ease of understanding, \textbf{we will focus on a simplified setting with group size $\omega=1$} (so we have $m=k,m^\prime=k^\prime, \cG_m = \cP_k, \bfg_j=\bfp_j$). This will cover the key ideas of certification algorithms, and we will generalize it to larger $\omega$ in Appendix~\ref{apx-cert-omega}. 

We will start with an overview and a warm-up example in Section~\ref{sec-cert-overview}, followed by certification algorithms for our keyword aggregation (Section~\ref{sec-cert-keyword}) and decoding aggregation (Section~\ref{sec-cert-decoding}), and conclude with remarks on the usage of certification procedures (Section~\ref{sec-cert-eval-remark}).

\subsection{Certification Overview}\label{sec-cert-overview}
\textbf{Certification objective.} Given a RAG query $\bfq$, the robustness certification procedure aims to determine the (largest) $\tau$ that satisfies $\tau$-robustness (Definition~\ref{dfn-certifiable}). Toward this objective, the certification procedure will analyze all possible \framework responses $\bfr$ when an attacker can arbitrarily inject $k^\prime$ malicious passages to the top-$k$ retrieval $\cP_k$. Let $\cR$ be the set of all possible \framework responses $\bfr$. We will show that, thanks to our \framework design, $\cR$ is a finite set. This allows us to measure the worst-case performance/robustness as $\tau=\min_{\bfr\in\cR}\mathbb(\texttt{M}(\bfr,\bfa))$, where $\bfa$ is the reference answer.

\textbf{Understanding LLM inputs.} To analyze all possible LLM outputs, we need to first understand possible LLM inputs (i.e., possible passages/groups). 
Recall that the attacker \textit{injects} $k^\prime$ passage to the retrieval result (Section~\ref{sec-prelim-attack}); these $k^\prime$ injected passages can only eject the \textit{bottom} $k^\prime$ benign passages $\{\bfp_{k-k^\prime+1},\dots,\bfp_{k}\}$ from the original retrieval result $\cP_k$. Therefore, the top $k-k^\prime$ benign passages $\{\bfp_1,\dots,\bfp_{k-k^\prime}\}$ remain unchanged in the corrupted retrieval set $\cP_k^\prime$, regardless of the content and ranking of injected passages.\footnote{Appendix~\ref{apx-modification} discusses passage modification that can eject/modify any $k^\prime$ passages instead of the bottom $k^\prime$.} Our robustness certification (with group size $\omega=1$) will be based on these top $k-k^\prime$ benign passages $\{\bfp_1,\dots,\bfp_{k-k^\prime}\}\coloneqq\cP_{k-k^\prime}$.

\textbf{Warm-up example: majority voting.} We can first use majority voting for \textit{classification} as a warm-up example to understand the certification workflow. 

\spnew{Consider a RAG pipeline is built for multiple-choice question-answering, where the output space is small and predefined. A simple aggregation strategy is to output the answer with the highest votes among $k$ isolated predictions (majority voting).} To analyze its robustness given $k-k^\prime$ benign passages we can get $k-k^\prime$ classification outputs and gather their voting counts. \textit{If the voting count difference between the winner and runner-up is larger than $k^\prime$}, the final response can only be the voting winner $\bfr^*$, regardless of the content and position of the $k^\prime$ corrupted passage groups. This is because the attacker can only increase the runner-up count by $k^\prime$ (using $k^\prime$ malicious passage groups), which is not enough for the runner-up to beat the winner. Therefore, we have $\cR=\{\bfr^*\}$ and thus we can compute $\tau=\texttt{M}(\bfr^*,\bfa)\in\{0,1\}$ in this case. On the other hand, if the voting count difference is not large enough, we have no guarantees on the output; the certification procedure should return $\tau=0$, i.e., zero robustness.

\subsection{Certifying Keyword Aggregation}\label{sec-cert-keyword}

\textbf{Intuition.} Similar to majority voting, we analyze the $k-k^\prime$ benign responses made from $k-k^\prime$ benign passages. We first extract keywords and get their counts as discussed in Section~\ref{sec-alg-keyword}. We next analyze which keywords might appear in the retained keyword set $\cW^*$ (Line~\ref{ln-keyword-filtering} of Algorithm~\ref{alg:keyword}). Intuitively, keywords with large counts will \textit{always} appear in $\cW^*$ while keywords with small counts can \textit{never} be in $\cW^*$. As a result, the attacker can only manipulate the appearance of keywords with ``medium'' counts. When the set of medium-count keywords is small (e.g., smaller than 15), we can enumerate all its possible subsets and generate all possible retained keyword set $\cW^*$ accordingly (by combining large-count and medium-count keywords). Finally, we compute all possible responses $\bfr$ from all possible $\cW^*$ and let them form a response set $\cR$---we have $\tau=\min_{\bfr\in\cR}\texttt{M}(\bfr,\bfa)$. 

\textbf{Certification algorithm.} We provide the certification pseudocode for keyword aggregation in Algorithm~\ref{alg:keyword-certify}. It aims to determine the $\tau$ value in $\tau$-certifiable robustness for a given query $\bfq$, benign passages $\cP_{k-k^\prime}$, the number of corrupted passages $k^\prime$, and given defense/attack settings. We state its correctness as follows. 

\begin{theorem}\label{thm1}
Given benign passages $\cP_{k}$, number of corrupted passages $k^\prime$, query $\bfq$, model $\llm$, group size $\omega=1$, filtering thresholds $\alpha,\beta$, and reference answer $\bfa$, Algorithm~\ref{alg:keyword-certify} can correctly return the $\tau$ value for $\tau$-certifiable robustness for the inference procedure \textsc{RRAG-keyword} discussed in Algorithm~\ref{alg:keyword}, i.e., $ \texttt{M}(\bfr,\bfa)\geq\tau, \forall \, \bfr\in \cR\coloneqq\{\textsc{RRAG-keyword}(\bfi,\bfq,\cPkp,\llm,\omega=1,\alpha,\beta)\,|\,\forall\, \cPkp \in \cA(\cP_k,k^\prime)\}$.
\end{theorem}

\begin{algorithm}[t]
\centering
\caption{Certification procedure for keyword aggregation}
\begin{algorithmic}[1]
\Require Benign passages $\cP_{k-k^\prime}$, number of corrupted passages $k^\prime$, query $\bfq$, model $\llm$, filtering thresholds $\alpha\in[0,1],\beta\in\mathbb{Z}^+$, reference answer $\bfa$.

\renewcommand{\algorithmicrequire}{\textbf{Instructions:}}
\Require  $\bfi_1$ = ``\stexttt{answer the query given retrieved passages, say `I don't know' if no relevant information found}''; $\bfi_2$ = ``\stexttt{answer the query using provided keywords}''

\State $\cC\gets\textsc{Counter}(),n\gets 0$ \label{ln-keyword-cert-init}
\For{$j\in\{1,2,\dots,k-k^\prime\}$}
    \State $\bfr_j\gets \llmgen(\bfi_1\cat\bfq\cat\bfp_j)$ 
    \If{$\text{``\stexttt{I don't know}''} \not\in\bfr_j$}
        \State $\cW_j\gets\textsc{GetUniqKeywords}(\bfr_j)$ 
        \State Update counter $\cC$ with $\cW_j$ 
        \State $n\gets n+1$ 
    \EndIf
\EndFor \label{ln-keyword-cert-getcount-end}
\State $\cR\gets\{\}$\label{ln-keyword-cert-emptyset}
\For{$k^\prime_\text{effective}\in\{0,1,\dots,k^\prime\}$}
\State $\mu^\prime\gets\min(\alpha \cdot (n+k^\prime_\text{effective}),\beta )$ \label{ln-keyword-cert-thres}
\State $\cW_A\gets\{\bfw|(\bfw,c)\in\cC,c\geq\mu^\prime\}$\label{ln-keyword-cert-WA}
\State $\cW_C\gets\{\bfw|(\bfw,c)\in\cC,\mu^\prime>c\geq\mu^\prime-k^\prime_\text{effective}\}$\label{ln-keyword-cert-WC}
\For{$\cW_C^\prime\in\mathbb{P}(\cW_C)$}\label{ln-keyword-cert-wcfor}
\State $\cW^\prime\gets\cW_A\bigcup\cW_C^\prime$\label{ln-keyword-cert-combine}
\State $\bfr\gets\llmgen(\bfi_2\cat\bfq\cat\textsc{Sorted}(\cW^\prime))$ \label{ln-keyword-cert-response}
\State $\cR\gets\cR\bigcup\{\bfr\}$ \label{ln-keyword-cert-union}
\EndFor\label{ln-keyword-cert-wcforend}
\EndFor
\State $\tau\gets\min_{\bfr\in\cR}\texttt{M}(\bfr,\bfa)$\label{ln-keyword-cert-min}
\State\Return $\tau$

\end{algorithmic}
\label{alg:keyword-certify}
\end{algorithm}

\begin{figure}
    \centering
    \includegraphics[width=\linewidth]{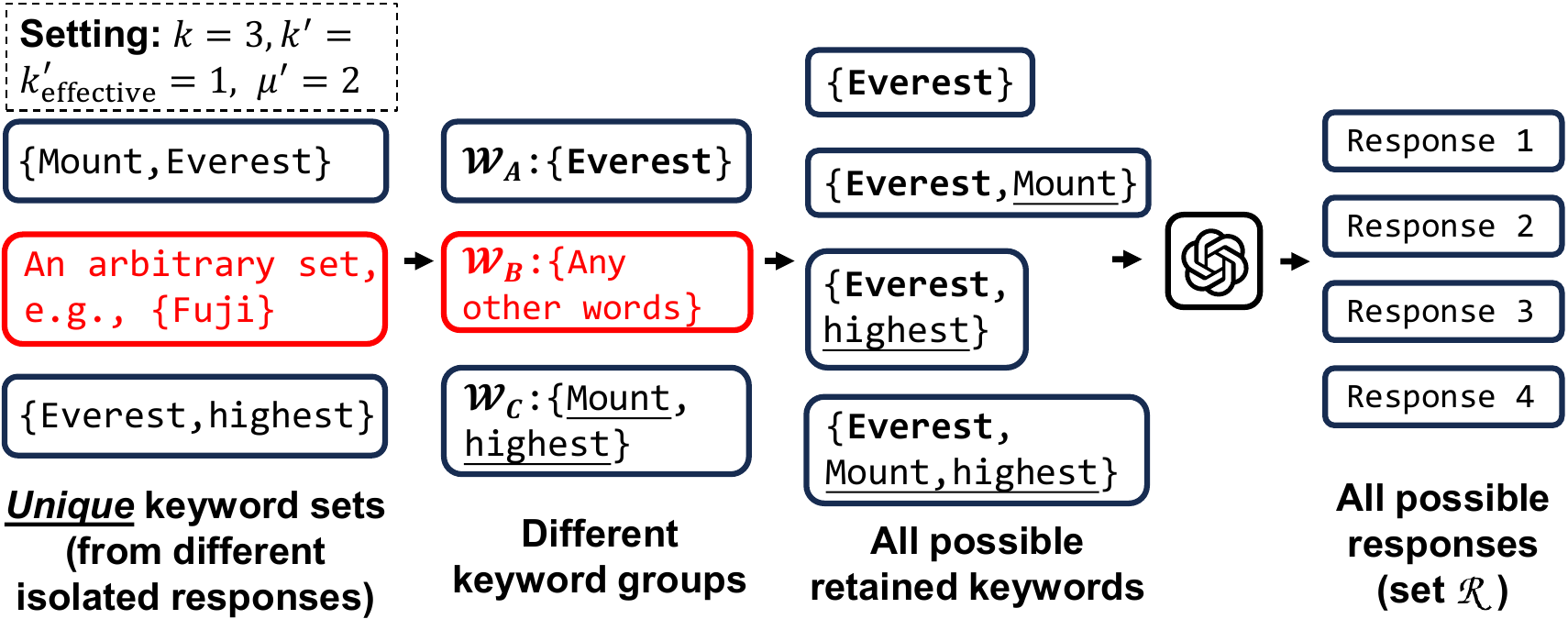}
    \caption{\textbf{Visual example of keyword certification.} One out of three passages is corrupted, and the attack can introduce any word to the corrupted keyword set. (1) Words from $\cW_A$ (with counts larger than or equal to $\mu^\prime=2$) will always be retained. (2) Words from $\cW_B$ (with counts smaller than $\mu^\prime-k^\prime_\text{effective}=1$) will always be filtered; therefore, malicious keywords like ``Fuji'' can never affect \framework output. (3) Words from $\cW_C$ with medium counts will be retained if the malicious keyword set contains the same words; therefore, the attacker has arbitrary control over their appearance. We can generate all possible retained keyword sets by enumerating the power set of $\cW_C$ (and combining them with the keyword set $\cW_A$). Given all possible retained keyword sets, we can prompt the LLM to generate all possible responses $\cR$.} 
    \label{fig-keyword-cert}
\end{figure}

\begin{proof}

    We prove the theorem by first explaining Algorithm~\ref{alg:keyword-certify}. We provide a toy example in Figure~\ref{fig-keyword-cert} to aid our discussion.
        
    First, the certification procedure extracts keywords and get their counts from the $k-k^\prime$ responses computed from benign passage groups (Lines~\ref{ln-keyword-cert-init}-\ref{ln-keyword-cert-getcount-end}). The keyword extraction process is identical to the inference algorithm discussed in Algorithm~\ref{alg:keyword}.

    Then, the certification procedure initializes an empty response set $\cR$ to collect all possible responses (Line~\ref{ln-keyword-cert-emptyset}). Since the attacker might introduce arbitrary numbers of non-abstained malicious responses (responses without ``\stexttt{I don't know}''), we denote this number as $k^\prime_\text{effective}$ and will enumerate all possible cases $k^\prime_\text{effective}\in\{0,1,\dots,k^\prime\}$. 
    
    For each $k^\prime_\text{effective}$, we first compute the corresponding threshold $\mu^\prime=\min(\alpha \cdot (n+k^\prime_\text{effective}),\beta)$, where $n$ is the number of non-abstained responses from $k-k^\prime$ benign passages (Line~\ref{ln-keyword-cert-thres}).
    Given the threshold $\mu^\prime$, we can divide all keywords into three groups (a toy example is in Figure~\ref{fig-keyword-cert}). 
    \begin{enumerate}
        \item The first group $\cW_A$ contains keywords with counts no smaller than $\mu^\prime$. Keywords from this group will always be in the retained keyword set $\cW^*$ because the injection attacker cannot decrease their counts.
        \item The second group $\cW_B$ contains keywords with counts smaller than $\mu^\prime-k^\prime_\text{effective}$. These keywords will never appear in the final keyword set $\cW^*$ because the attacker can only increase their counts by $k^\prime_\text{effective}$.
        \item The third group $\cW_C$ contains keywords whose counts are within $[\mu^\prime-k^\prime_\text{effective}, \mu^\prime)$. The attacker can arbitrarily decide if these keywords will appear in the retained keyword set.
    \end{enumerate}   

    We then generate keyword sets $\cW_A$ and $\cW_C$ accordingly (Lines~\ref{ln-keyword-cert-WA}-\ref{ln-keyword-cert-WC}). Note that we do not need $\cW_B$ for certification as it will not be part of the retained keyword set. Next, we enumerate all possible keyword sets from the power set $\cW^\prime_C\in\mathbb{P}(\cW_c)$. For each $\cW^\prime_C$, we generate retained keyword set $\cW^\prime=\cW_A\bigcup\cW_C^\prime$ (Line~\ref{ln-keyword-cert-combine}), obtain the corresponding response $\bfr= \llmgen(\bfi_2\cat\bfq\cat\textsc{Sorted}(\cW^\prime))$ (Line~\ref{ln-keyword-cert-response}), and add this response to the response set (Line~\ref{ln-keyword-cert-union}).

    After we enumerate all possible $k^\prime_\text{effective}$ and all possible retained keyword set $\cW^\prime$. The set $\cR$ contains all possible LLM responses. We call the evaluation metric function $\texttt{M}(\cdot)$ and get the lowest score as the certified $\tau$ value (Line~\ref{ln-keyword-cert-min}).

    In summary, the certification procedure has considered all possible responses and returns the lowest evaluation metric score. Therefore, the returned value is the correct $\tau$ value for certifiable robustness.
\end{proof}
\textbf{Implementation details.} When the keyword power set $\mathbb{P}(\cW_C)$ is too large and costly to enumerate (e.g., $2^{15}$ with $|\cW_C|>15$), we conservatively consider the certification fails and return $\tau=0$, i.e., zero-certifiable robustness. 

\begin{algorithm}[t]
    \caption{Certification procedure for decoding aggregation}
    \begin{algorithmic}[1]

    \Require Benign passages $\cP_{k-k^\prime}$, number of corrupted passages $k^\prime$, query $\bfq$, model $\llm$, threshold $\gamma$, probability threshold $\eta$, max number of new tokens $T_\text{max}$, reference answer $\bfa$.
    
    \renewcommand{\algorithmicrequire}{\textbf{Instruction:}}
    \Require $\bfi$ = ``\stexttt{answer the query given retrieved passages, say `I don't know' if no relevant information found}''

    \State $\cR \gets \{\},\cX\gets\textsc{Stack}(\{\text{``''}\})$ \label{ln-decoding-cert-init}

    \State $\cJ\gets \{j|\Pr_\llm[\text{``\stexttt{I don't know}''}|\bfi\cat\bfq\cat\bfp_j]<\gamma,\bfp_j\in\cP_{k-k^\prime}\}$  \label{ln-decoding-cert-filtering}
    \While{$\cX$ is not empty}
    \State $\hat{\bfr}\gets\cX.\textsc{Pop}()$ \label{ln-decoding-cert-pop}
    \If{$\textsc{Len}(\hat{\bfr})\geq T_\text{max}$}
    \State $\cR\gets\cR\bigcup\{\hat{\bfr}\}$\label{ln-decoding-cert-add}
    \State \textbf{continue}
    \EndIf

        \State $\hat{\bfv}\gets \textsc{Vec-Sum}(\{\bfv_j|\bfv_j=\llmconf(\bfi\cat\bfq\cat\bfp_j\cat\bfr^*), j\in\cJ\})$  \label{ln-decoding-cert-aggregate}
        \State $(\bft_a,A),(\bft_b,B)\gets\textsc{Top2Tokens}(\hat{\bfv})$  \label{ln-decoding-cert-top2}
        \State $\bft_\text{nor}\gets\llmtoken(\text{``\stexttt{answer query}''}\cat\bfq\cat\hat{\bfr})$\label{ln-decoding-cert-zero}
        \If{$A-B> \eta+k^\prime$}\label{ln-decoding-cert-case1cond}
        \State $\cX.\textsc{Push}(\hat{\bfr}\cat\bft_a)$ \label{ln-decoding-cert-case1}
        \ElsIf{ $( \eta+k^\prime\geq A-B>| \eta-k^\prime|)$ }\label{ln-decoding-cert-case2cond}
        \State $\cX.\textsc{Push}(\hat{\bfr}\cat\bft_a);\cX.\textsc{Push}(\hat{\bfr}\cat\bft_\text{nor})$ \label{ln-decoding-cert-case2}
        \ElsIf{ $( \eta-k^\prime\geq A-B>0)$ }\label{ln-decoding-cert-case3cond}
        \State $\cX.\textsc{Push}(\hat{\bfr}\cat\bft_\text{nor})$\label{ln-decoding-cert-case3}
        \Else \label{ln-decoding-cert-case4cond}
         \State\Return $0$\label{ln-decoding-cert-case4}
        \EndIf\label{ln-decoding-cert-4caseend}
    \EndWhile
\State $\tau\gets\min_{\bfr\in\cR}\texttt{M}(\bfr,\bfa)$\label{ln-decoding-cert-tau}
\State\Return$\tau$
    \end{algorithmic}
    \label{alg:decoding-certify}
    \end{algorithm}

\subsection{Certifying Decoding Aggregation}\label{sec-cert-decoding} 

\textbf{Intuition.} We aim to analyze all possible next-token predictions at every decoding step. Given a partial response at a certain decoding step, we first compute next-token probability vectors predicted on $k-k^\prime$ benign passage groups and calculate the probability sum of each token. Next, we identify the top-2 tokens with the largest probability sums and compute their probability difference as $\delta$. We will use this $\delta$ value to analyze possible next-token predictions. Intuitively, a large $\delta$ always leads to the top-1 token being predicted; a medium $\delta$ allows for predictions of either the top-1 token or the no-retrieval token; when $\delta$ is small, the prediction can be any malicious token introduced by the attacker. We start our certification with an empty string and track all possible next-token predictions (and partial responses) at different decoding steps. If $\delta$ is never ``small'' when we finish decoding all possible responses; we can obtain a finite set of all possible responses $\cR$---we have $\tau=\min_{\bfr\in\cR}\texttt{M}(\bfr,\bfa)$. 

\textbf{Certification algorithm.} We present the pseudocode in Algorithm~\ref{alg:decoding-certify}. It aims to determine the $\tau$ value in $\tau$-certifiable robustness for a given query $\bfq$, benign passages $\cP_{k-k^\prime}$, the number of corrupted passages $k^\prime$, and given defense/attack settings. We state its correctness as follows. 

\begin{theorem}\label{thm2}
Given benign passages $\cP_k$, number of corrupted passages $k^\prime$, query $\bfq$, model $\llm$, group size $\omega=1$, filtering thresholds $\gamma$, probability threshold $ \eta$, max number of new tokens $T_\text{max}$, and reference answer $\bfa$, Algorithm~\ref{alg:decoding-certify} can correctly return the $\tau$ value for $\tau$-certifiable robustness for the inference procedure \textsc{RRAG-decoding} discussed in Algorithm~\ref{alg:decoding}, i.e., $ \texttt{M}(\bfr,\bfa)\geq\tau, \forall \, \bfr\in \cR\coloneqq\{\textsc{RRAG-decoding}(\bfi,\bfq,\cPkp,\llm,\omega=1,\gamma, \eta,T_\text{max})\,|\,\forall\, \cPkp \in \cA(\cP_k,k^\prime)\}$.
\end{theorem}

\begin{proof}
    We prove the theorem by first explaining Algorithm~\ref{alg:decoding-certify}. 
    
    First, we initialize an empty response set $\cR$ to hold all possible responses and a stack $\cX$ with \textit{an empty string} to track possible \textit{partial} responses (Line~\ref{ln-decoding-cert-init}). Then, we get the indices of benign passage groups that are unlikely to output ``\stexttt{I don't know}'' (Line~\ref{ln-decoding-cert-filtering}). 
    We will repeat the following robustness analysis until the stack is empty. At each analysis step, we pop a partial response $\hat{\bfr}$ from the stack $\cX$ (Line~\ref{ln-decoding-cert-pop}). If it has reached the maximum number of generated tokens (or ends with an EOS token), we add this response $\hat{\bfr}$ to the response set $\cR$ (Line~\ref{ln-decoding-cert-add}). Otherwise, we get the probability sum vector $\hat{\bfv}$ from benign passages (Line~\ref{ln-decoding-cert-aggregate}) and its top-2 tokens $\bft_a,\bft_b$ and their probability sums $A,B$ (Line~\ref{ln-decoding-cert-top2}). We also get the no-retrieval prediction token as $\bft_\text{nor}$ (Line~\ref{ln-decoding-cert-zero}). 

    Next, we analyze all possible next-token predictions at this decoding step. Our discussions are based on the probability gap between $A$ and $B$, i.e., $A-B$. We can have the following three lemmas. We leave their proofs in Appendix~\ref{apx-certification}.

\begin{lemma}\label{lemma1}
    If $A-B> \eta+k^\prime$ is true, the algorithm will always predict the top-1 $\bft_a$.
\end{lemma}

\begin{lemma}\label{lemma2}
    If $ \eta+k^\prime \geq  A-B > | \eta-k^\prime|$ is true, the algorithm might predict the top-1 token $\bft_a$ or the no-retrieval token $\bft_\text{nor}$, but not any other token.
\end{lemma}

\begin{lemma}\label{lemma3}
    If $ \eta-k^\prime \geq A-B>0$ is true, the algorithm will always predict the no-retrieval token $\bft_\text{nor}$.
\end{lemma}

\noindent With these three lemmas, we can go back to the certification procedure in Algorithm~\ref{alg:decoding-certify}. We have four cases in total.

\begin{enumerate}
    \item \textit{Case 1}: $A-B> \eta+k^\prime$ (Line~\ref{ln-decoding-cert-case1cond}). Lemma~\ref{lemma1} ensures that the next token is the top-1 token $\bft_a$; thus, we push $\hat{\bfr}\cat\bft_a$ to the stack $\cX$ (Line~\ref{ln-decoding-cert-case1}).
    \item \textit{Case 2}: $ \eta+k^\prime\geq A-B> | \eta-k^\prime| $ (Line~\ref{ln-decoding-cert-case2cond}). Lemma~\ref{lemma2} ensures that the next token is either top-1 token $\bft_a$ or no-retrieval token $\bft_\text{nor}$, which is under the attacker's control; we push both $\hat{\bfr}\cat\bft_a$ and $\hat{\bfr}\cat\bft_\text{nor}$ to $\cX$ (Line~\ref{ln-decoding-cert-case2}).
    \item  \textit{Case 3}: $  \eta-k^\prime \geq A - B > 0$ (Line~\ref{ln-decoding-cert-case3cond}). Lemma~\ref{lemma3} ensures that the next token is the no-retrieval token $\bft_\text{nor}$; thus, we push $\hat{\bfr}\cat\bft_\text{nor}$ to $\cX$ (Line~\ref{ln-decoding-cert-case3}).
    \item \textit{Case 4}: other cases. We cannot claim any robustness about the next-token prediction: the response set becomes intractable, and the robustness certification fails. Therefore, the algorithm aborts and returns $\tau=0$, i.e., zero-certifiable robustness (Line~\ref{ln-decoding-cert-case4}).
\end{enumerate}

Finally, if the response set $\cR$ is still tractable (no \textit{Case 4} happens) when the stack $\cX$ becomes empty, we return $\tau$ as the worst evaluation score $\min_{\bfr\in\cR}\texttt{M}(\bfr,\bfa)$ (Line~\ref{ln-decoding-cert-tau}).

In summary, Algorithm~\ref{alg:decoding-certify} considers all possible responses and returns the lowest evaluation metric score---the returned value is the correct $\tau$ value for certifiable robustness.    
\end{proof}

\begin{figure}
    \centering
    \includegraphics[width=\linewidth]{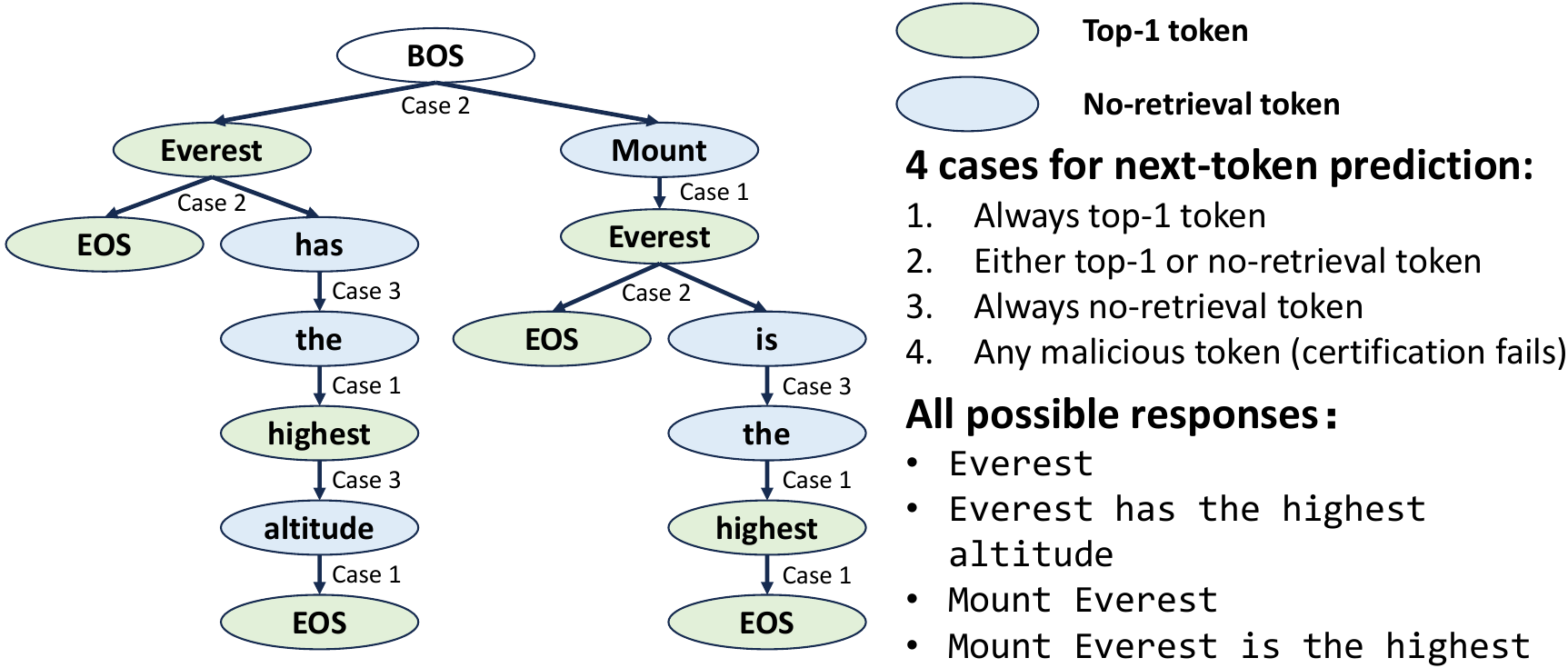}
    \caption{\textbf{Visual example of decoding certification.} The certification process can be viewed as a binary tree generation process, where each token corresponds to a tree node. We start with the BOS token (root node) and analyze the next-token prediction at each decoding step. If we hit Case 2, we branch out with two nodes (top-1 token and no-retrieval token); if we hit Case 1 or Case 3, we append the top-1 or no-retrieval token accordingly; if we hit Case 4, the certification fails, and algorithms aborts with $\tau=0$ (zero certifiable robustness).  If we finish the tree generation (end with EOS tokens or reach the maximum number of newly generated tokens), each root-to-leaf path corresponds to one possible LLM response. We compute $\tau$ as the lowest evaluation score from all these responses.}
    \label{fig-decoding-cert}
\end{figure}

We note that the entire certification process can be viewed as a binary tree generation, where each next-token prediction is a tree node. We provide a toy example in Figure~\ref{fig-decoding-cert} (see figure caption for more details).

\textbf{Implementation details.} The number of possible responses $|\cR|$ can sometimes become very large ($>10^3$) when \textit{Case 2} happens frequently. In \textit{a small subset} of our experiments, we sample a random subset $\hat{\cR}$ (of size 100) from the large response set $\cR$ and approximate the $\tau$ value as $\hat{\tau}=\min_{\bfr\in\hat{\cR}}\texttt{M}(\bfr,\bfa)$, due to the high financial and computational cost when using LLM-as-a-judge as evaluator $\texttt{M}(\cdot)$. These approximated robustness numbers will be marked with $^\ddag$ in Section~\ref{sec:exp}.


\subsection{Remark: Certifiable Evaluation} \label{sec-cert-eval-remark}
So far, we have discussed certification algorithms for different text aggregation techniques. The certification algorithms allow us to analyze the response set $\cR$ to determine the $\tau$ value of $\tau$-certifiable robustness for a given query $\bfq$ and its reference answer $\bfa$. In our evaluation, we gather a dataset of queries and answers $(\bfq,\bfa)$, calculate the $\tau$ value for each query, and take the averaged $\tau$ across different queries as a \textit{certifiable} evaluation metric of robustness. The evaluated robustness numbers are agnostic to attack algorithms and hold for strong adaptive attacks. 

We note that the \textit{certification} algorithms discussed in this section (Algorithms~\ref{alg:keyword-certify} and \ref{alg:decoding-certify}) are different from the \textit{inference} algorithms (Algorithms~\ref{alg:keyword} and \ref{alg:decoding}) discussed in Section~\ref{sec:algorithm}. The inference algorithms are the defense algorithms we will deploy in the wild; they aim to generate accurate responses from benign or corrupted retrieval. In contrast, the certification algorithms are designed to \textit{certifiably and provably evaluate} the robustness of inference algorithms (i.e., computing correct $\tau$); they operate on benign passages, require the reference answer $\bfa$ (to compute quality scores), and can be computationally expensive (to reason about all possible $\bfr\in\cR$).

\new{Finally, our robustness certification is compatible with any evaluation function $\texttt{M}$. We generate the set $\cR$ of all possible responses, and $\texttt{M}$ is used solely to compute a \textit{quantitative} robustness value $\min_{\bfr \in \cR} \texttt{M}(\bfr,\bfa)$. For long-form text generation, although LLM-as-a-judge is not a perfect evaluator, we will adopt it as it is the most effective automatic option. Nevertheless, our certification remains compatible with any alternative or improved evaluator $\texttt{M}$ (e.g., human evaluation).}

\section{Evaluation}
\label{sec:exp}

\definecolor{Gray}{gray}{0.9}
\definecolor{shadecolor}{rgb}{0.9,0.9,0.9}
	
\begin{table*}[t]
\centering
\caption{Certifiable robustness and benign performance of \framework ($k=10,k^\prime=1$).} %
\label{tab:main_cr}

   \vspace{-1em}

{
\begin{threeparttable}

\begin{tabular}{llcccccccc}

\toprule
Task& \multicolumn{1}{l}{\multirow{2}{*}{Model/ }}&\multicolumn{2}{c}{Multiple-choice QA} &  \multicolumn{4}{c}{Short-answer QA} & \multicolumn{2}{c}{Long-form generation}     \\
Dataset&\multicolumn{1}{l}{\multirow{2}{*}{Defense}} & \multicolumn{2}{c}{RQA-MC}   &  \multicolumn{2}{c}{RQA}  & \multicolumn{2}{c}{NQ} &  \multicolumn{2}{c}{Bio}   \\ 
\multicolumn{1}{l}{LLM} & & {(bacc)} & (cacc) & (bacc) & (cacc)  & (bacc) & (cacc) & (bllmj) & (cllmj)  \\
\midrule
\multirow{5}{*}{Mistral-I$_{\textsc{7b}}$}&No RAG & 9.0 &--  & 8.0 & --& 30.0 & -- & 59.4&--   \\
\cdashline{2-10}\noalign{\vspace{1.5pt}} 

& Vanilla & 80.0 & --  & 69.0  & --  & 61.0  & --   &  78.4 & --         \\
\cdashline{2-10}\noalign{\vspace{1.5pt}} 
& Keyword  &  &   & 71.0  & 38.0  & 61.0  & 26.0  & 64.8   & 46.6  \\ 
\cdashline{2-2}\cdashline{5-10}\noalign{\vspace{1.5pt}} 

& Decoding$_c$ &  &  &    &   &  &     & 71.2  &  45.6\tnote{$\ddag$}  \\
& Decoding$_r$  &  \multirow{-3}{*}{81.0\tnote{$\dag$}}&\multirow{-3}{*}{71.0\tnote{$\dag$}}  & \multirow{-2}{*}{62.0}  & \multirow{-2}{*}{37.0} &  \multirow{-2}{*}{62.0}  &  \multirow{-2}{*}{29.0} & 63.4  &  51.2  \\
\midrule
\multirow{5}{*}{Llama2-C$_{\textsc{7b}}$  }
&No RAG  & 21.0 & -- & 2.0 & -- & 10.0 &--  &   19.6 &--   \\
\cdashline{2-10}\noalign{\vspace{1.5pt}} 

& Vanilla & 79.0 & --  & 61.0  & --   & 57.0  & --   &  71.8 & --     \\
\cdashline{2-10}\noalign{\vspace{1.5pt}} 
& Keyword  &   &   & 64.0  & 34.0  & 56.0  & 31.0  & 62.2  &  46.4     \\ 
\cdashline{2-2}\cdashline{5-10}\noalign{\vspace{1.5pt}} 
& Decoding$_c$   & &  &    &   &   &   & 70.6  &  38.8\tnote{$\ddag$}   \\
&  Decoding$_r$  & \multirow{-3}{*}{78.0\tnote{$\dag$}} &  \multirow{-3}{*}{69.0\tnote{$\dag$}}& \multirow{-2}{*}{61.0}  & \multirow{-2}{*}{31.0}  & \multirow{-2}{*}{53.0} &  \multirow{-2}{*}{36.0} & 62.4  &  41.6   \\

\bottomrule
\end{tabular}
 \begin{tablenotes}
 \item (bacc): accuracy; (cacc): certifiable accuracy; (bllmj): LLM-judge score; (cllmj): certifiable LLM-judge score
 \item[$\dag$] \spnew{For multiple-choice QA, we only have one defense instance that uses majority voting for secure aggregation.}
 \item[$\ddag$] Approximated via subsampling, as discussed at the end of Section~\ref{sec-cert-decoding}. 
 \end{tablenotes}
\end{threeparttable}}
\end{table*}

We present the experimental setup in Section~\ref{subsec:evalsetup}, main results of certifiable robustness in Section~\ref{subsec:evalmain}, empirical attack experiments in Section~\ref{subsec:evalempirical}, and more detailed analysis of \framework in Section~\ref{subsec:evalabl}.

\subsection{Experiment Setup}
\label{subsec:evalsetup}
In this section, we discuss our experiment setup; we provide more details in Appendix~\ref{apx-details-implementation}. 

\textbf{Datasets.} We experiment with four datasets: \textbf{RealtimeQA-MC (RQA-MC)}~\citep{kasai2024realtime} for \textit{multiple-choice open-domain QA} (a classification task), \textbf{RealtimeQA (RQA)}~\citep{kasai2024realtime} and \textbf{Natural Questions (NQ)}~\citep{Kwiatkowski2019NaturalQA} for \textit{short-answer open-domain question-answering (QA)}, and the \textbf{Biography generation dataset (Bio)}~\citep{min-etal-2023-factscore} for \textit{free-form long text generation}, which generates a few sentences or paragraphs. We sample 100 queries (50 for Bio) from each dataset for experiments (as certification can be computationally expensive). For each query, we use Google Search to retrieve passages. This is a popular experiment setting~\citep{kasai2024realtime,yan2024corrective,vu2023freshllms} and mimics a real-world scenario where malicious webpages are retrieved.  We note that our \framework design is agnostic to the choice of retriever.

\textbf{LLM and RAG settings.} We evaluate \framework with three LLMs: Mistral-7B-Instruct~\citep{jiang2023mistral}, Llama2-7B-Chat~\citep{touvron2023llama}, and GPT-3.5-turbo (deferred to Appendix~\ref{apx-experiment-abl}).
We use in-context learning to guide LLMs to follow instructions. We use the top 10 retrieved passages for generation by default. We use greedy decoding for a deterministic evaluation of certifiable robustness. 

\textbf{\framework setup.} 
We evaluate \framework with two aggregation methods: secure keyword aggregation \textbf{(Keyword)} and secure decoding aggregation \textbf{(Decoding)}. By default, we set $k=10,\omega=1, \gamma = 0.99$. For multiple-choice QA (classification), we reduce \framework to \textit{majority voting}, which was discussed in Section~\ref{sec-certification}. For short-answer QA, we further set $\alpha = 0.2,\beta=3, \eta = 0$. For long-form generation, we set $\alpha = 0.4,\beta=4$ and include two secure decoding instances: one optimized for benign performance ($\eta = 1$), denoted by \textbf{Decoding$_c$}, and another for robustness ($\eta = 4$), denoted by \textbf{Decoding$_r$}. \spnew{We chose the default parameters using a held-out validation set and will analyze the impact of different parameters in Section~\ref{subsec:evalabl} and Appendix~\ref{apx-experiment-abl}.} 

\begin{table*}[t]
\centering
    \caption{Empirical robustness of \framework ($k=10,k^\prime=1$) against PIA and Poison attacks. }
    \label{tab:main_em}
    \vspace{-1em}
{
\begin{threeparttable}
\begin{tabular}{llcccccccc}\toprule
Task&   &   \multicolumn{4}{c}{Short-form open-domain QA} & \multicolumn{2}{c}{Long-form generation}     \\
Dataset&    \multirow{2}{*}{Model/}&    \multicolumn{2}{c}{RQA}  & \multicolumn{2}{c}{NQ} &  \multicolumn{2}{c}{Bio}   \\ 
Attack&  \multirow{2}{*}{Defense} & PIA   & Poison   & PIA   & Poison & PIA   & Poison  \\ 
LLM &  & racc\gua / asr\rda & racc\gua / asr\rda & racc\gua / asr\rda & racc\gua / asr\rda & rllmj\gua / asr\rda & rllmj\gua  / asr\rda \\
\midrule \noalign{\smallskip}
\multirow{3}{*}{Mistral-I$_{\textsc{7b}}$}
& Vanilla & 5.0 / 66.0  &  16.0 / 80.0 &  8.0 / 85.0  & 41.0 / 37.0 &  29.0  / 100   &  56.0  / 86.0   \\
& Keyword  & \textbf{72.0} / 15.0  & \textbf{72.0} / 15.0   &  \textbf{62.0} / 11.0  & \textbf{64.0} / 12.0  &   64.8  / \textbf{0.0}    &  61.6  / \textbf{0.0}    \\ 
& Decoding$_c$   & 57.0 / \textbf{5.0}  & 56.0 / 11.0   &  65.0 / \textbf{7.0}   & 63.0 / 7.0  &  \textbf{69.8}  / \textbf{0.0} &  \textbf{71.0}   / \textbf{0.0}  \\
\noalign{\smallskip} \hdashline \noalign{\smallskip}
 \multirow{3}{*}{Llama2-C$_{\textsc{7b}}$}
& Vanilla &  1.0 / 97.0  & 9.0 / 76.0   &  2.0 / 93.0  &  33.0 / 38.0  & 18.2  / 98.0   &    42.4  / 44.0   \\
& Keyword  &  64.0 / 12.0  & 64.0 / 11.0   &  55.0 / 10.0   &  55.0 / 9.0  &   59.2  / \textbf{0.0} &  63.4  / \textbf{0.0}      \\ 
& Decoding$_c$   &  59.0 / 5.0  & 60.0 / \textbf{3.0}   &  51.0 / 6.0  &  51.0 / \textbf{5.0}  &  67.6  / \textbf{0.0} &    67.8  / \textbf{0.0}  \\
\bottomrule
\end{tabular}
 \begin{tablenotes}
 \item (racc): empirical robust accuracy; (rllmj): empirical robust LLM-judge score;  (asr): targeted attack success rate 
 \end{tablenotes}
\end{threeparttable}
}
\end{table*}

\begin{figure*}
\centering
\begin{minipage}[t]{0.25\linewidth}
\includegraphics[width=\linewidth]{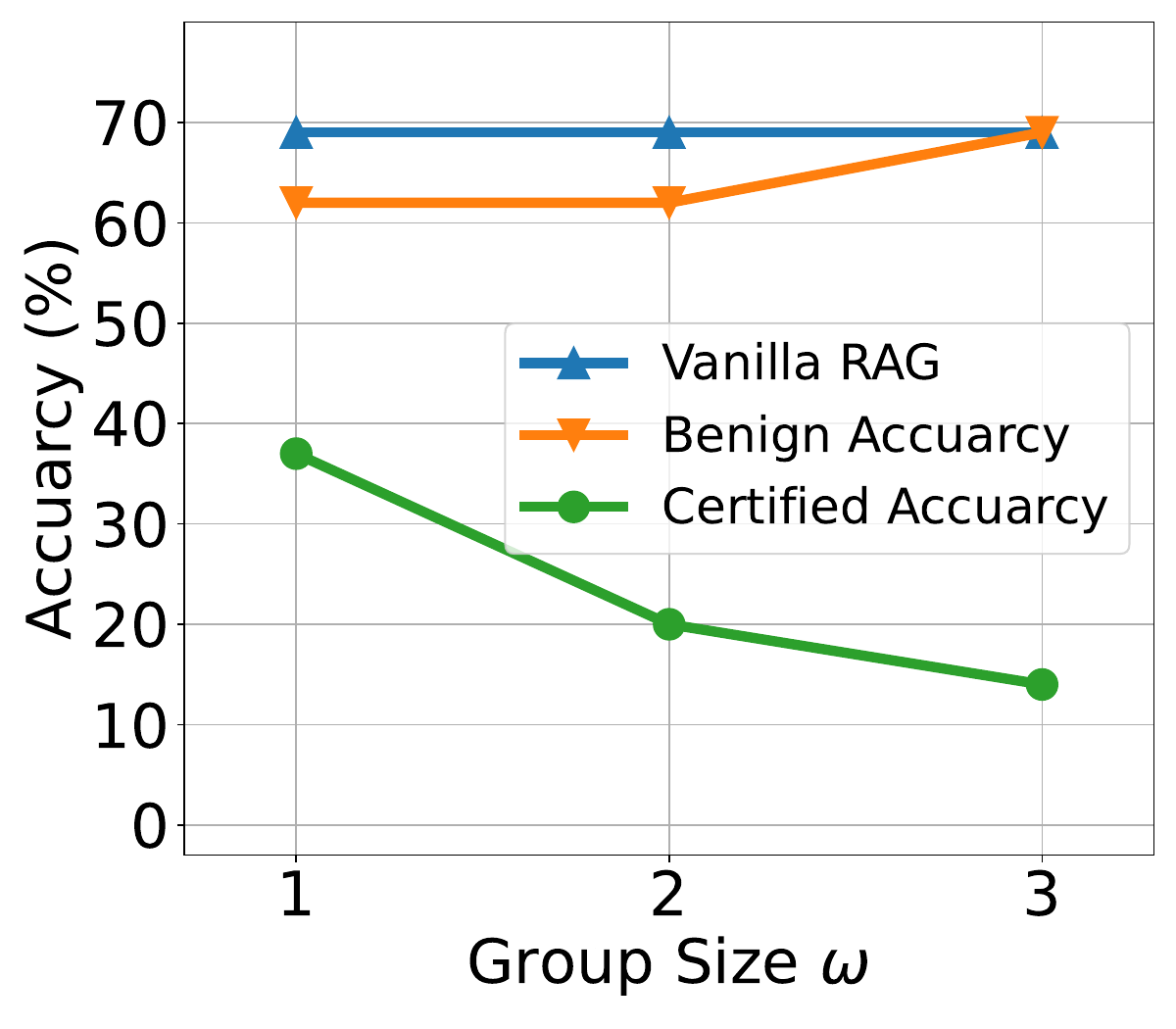}
 \vspace{-2.5em}
\caption{Effect of passage group size $\omega$ (RQA) }
\label{fig-gs}
\end{minipage}%
\quad
\begin{minipage}[t]{0.25\linewidth}
\includegraphics[width=\linewidth]{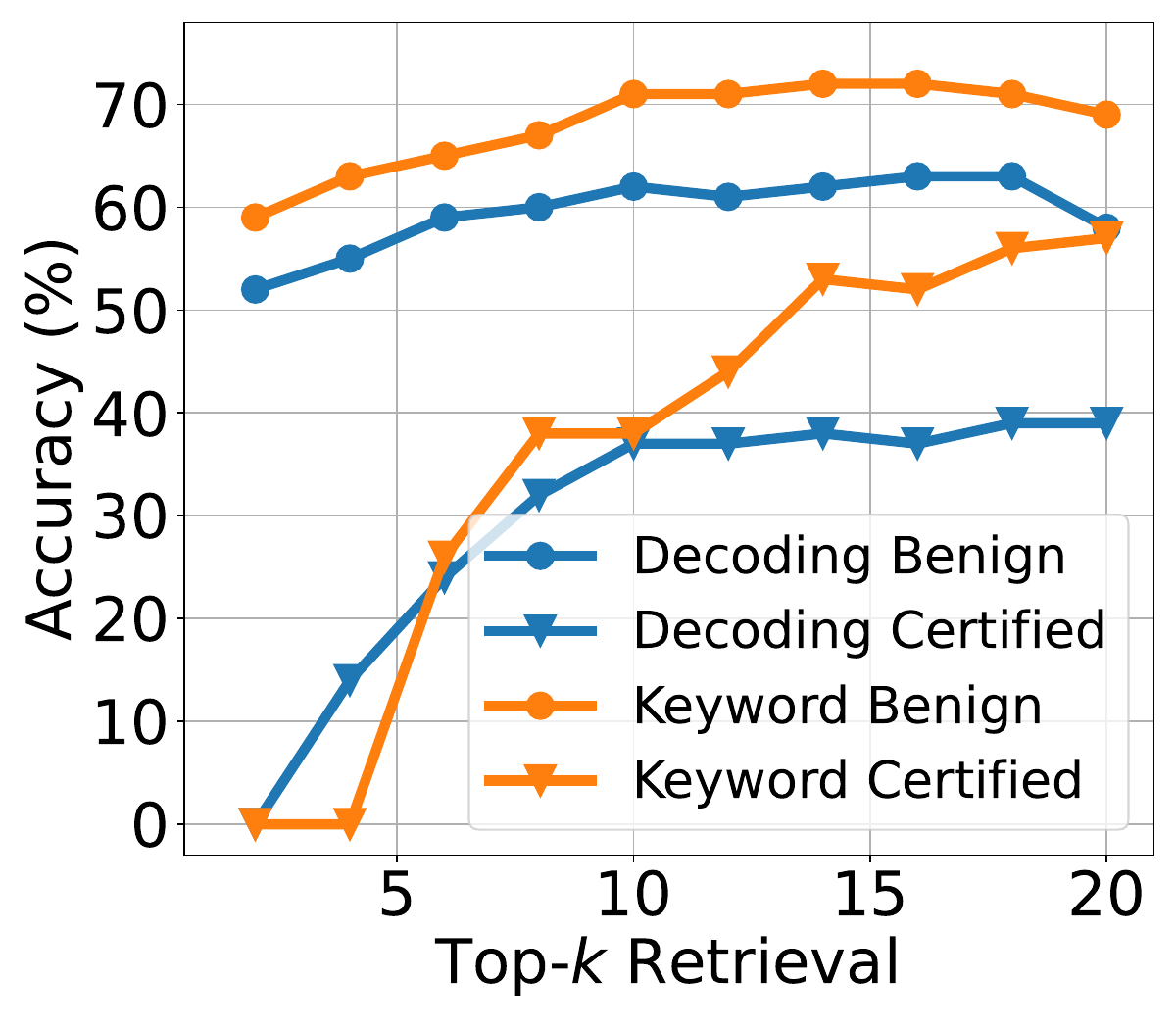}
 \vspace{-2.5em}
\caption{Effect of number of retrieved passages $k$ (RQA)}\label{fig-topk-rqa-mistral} 
\end{minipage}%
\quad
\begin{minipage}[t]{0.25\linewidth}
\includegraphics[width=\linewidth]{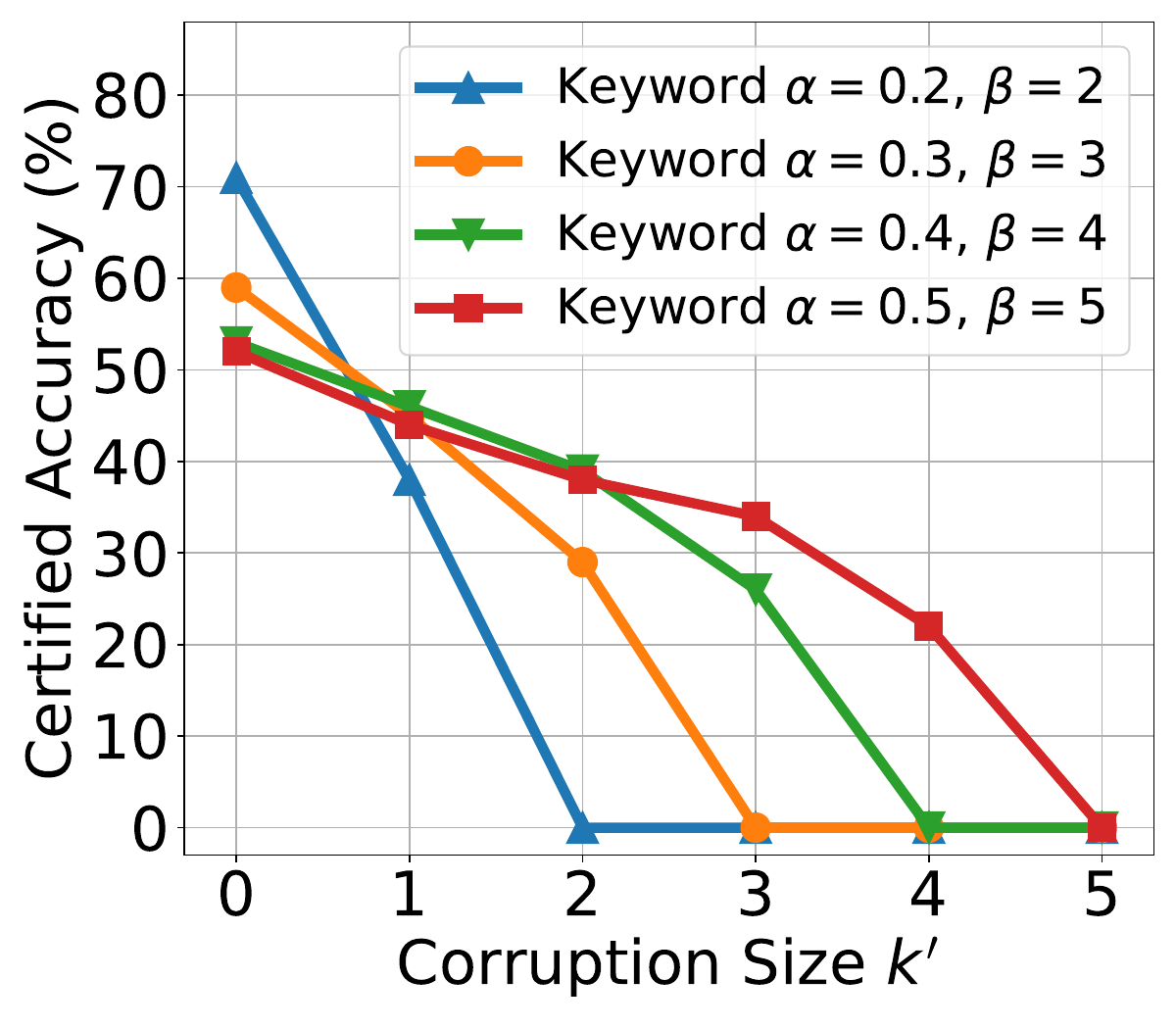}
 \vspace{-2.5em}
\caption{Effect of corruption size $k^\prime$ and keyword filtering thresholds $\alpha,\beta$ (RQA)} \label{fig-keyword-rqa-mistral}
\end{minipage}%
\quad

\end{figure*}
\textbf{Evaluation metrics.} \textit{For QA tasks}, we use the reference answer $\bfa$ to evaluate the correctness of the response. The evaluator $\texttt{M}(\cdot)$ returns a score of 1 when the reference answer $\bfa$ appears in the response $\bfr$, and outputs 0 otherwise. For benign performance evaluation (without any attack), we report the averaged evaluation scores on different queries as benign accuracy \textbf{(bacc)}. For certifiable robustness evaluation, we compute the $\tau$ values for different queries and report the averaged $\tau$ as the certifiable accuracy \textbf{(cacc)}.
\textit{For long-form bio generation}, we generate a reference response $\bfa$ by prompting GPT-4 with the person's Wikipedia document. We then use GPT-3.5 to build an LLM-as-a-judge evaluator~\citep{zheng2023judging} and rate responses with scores ranging from 0 to 100 \textbf{(bllmj)}. For robustness evaluation, we report the $\tau$ values as certifiable LLM-judge scores \textbf{(cllmj)}. 

\subsection{Main Evaluation of Certifiable Robustness}\label{subsec:evalmain}

In Table~\ref{tab:main_cr}, we report the certifiable robustness and benign performance of \framework with $k = 10$ passages, isolated by a group size of $\omega=1$, against $k^\prime = 1$ malicious passage. We also report performance for LLMs without retrieval \textbf{(no RAG)} and vanilla RAG with no defense \textbf{(vanilla)}.

\textbf{\framework achieves substantial certifiable robustness across different tasks and models.} As shown in Table~\ref{tab:main_cr}, \framework achieves 69.0--71.0\% certifiable robust accuracy for RQA-MC, 31.0--38.0\% for RQA, 26.0--36.0\% for NQ, and 38.8--51.2 certifiable LLM-judge score for the bio generation task. A certifiable accuracy of 71.0\% means that for 71.0\% of RAG queries, \framework's response will always be correct, even when the attacker knows everything about our framework and can inject anything into one retrieved passage. \framework achieves the \textit{first} certifiable robustness against any (adaptive) retrieval corruption attack strategy.

\textbf{\framework maintains high benign performance.} \spnew{Large drops in benign performance have been a significant challenge for certifiably robust defenses. For example, popular certifiably robust classifiers~\cite{xiang2021patchguard,crown,cohen2019certified} can incur over 10-20\% benign accuracy drop on the ImageNet dataset~\cite{imagenet}. In contrast, \framework can maintain high benign performance while providing substantial certifiable robustness.} For QA tasks, \framework has a minimal impact on benign performance in most cases (compared to vanilla RAG). The only exception is Mistral with secure decoding on RQA (a 7\% drop). However, we note that we can minimize this drop with a larger group size $\omega$---Figure~\ref{fig-gs} demonstrates that we can reduce the benign accuracy drop from 7\% to 0\% by setting the group size $\omega=3$.
For the long-form bio generation task, the benign performance drops can be as small as 1.2\% (Llama with {Decoding$_c$}); the drops are within 10\% in most other cases.  Finally, we note that \framework performs much better than generation without retrieval (no RAG)---\framework allows us to benefit from retrieval with robustness guarantees.

\subsection{\framework against Empirical Attacks}\label{subsec:evalempirical}

In Table \ref{tab:main_em}, we analyze the empirical robustness of \framework against prompt injection \textbf{(PIA)}~\citep{greshake2023not} and data poisoning \textbf{(Poison)}~\citep{zou2024poisonedrag}. We present the empirical robust accuracy \textbf{(racc)} or robust LLM-judge score \textbf{(rllmj)} against two attacks, as well as the targeted attack success rate \textbf{(asr)}, defined as the percentage of queries for which LLM returns the malicious responses chosen by the attacker.  
we observe that vanilla RAG pipelines are vulnerable to prompt injection and data poisoning attacks---PIA can have a 90+\% attack success rate and degrade the performance below 20\%. In contrast, \textbf{\framework achieves substantial empirical robustness}: the attack success rates are $\leq$ 15\% in all cases, with high robust accuracy/score.

\subsection{Detailed Analysis of \framework}  \label{subsec:evalabl}

In this section, we use Mistral-7B-Instruct to analyze its defense performance with different parameters.  
Appendix \ref{apx-experiment-abl} provides additional analyses for more models and datasets. 

\textbf{Impact of passage group size $\omega$.} Figure~\ref{fig-gs} analyzes group sizes $\omega$. As the group size $\omega$ increases, the benign performance generally improves, but the certifiable robustness gradually decreases.  The parameter $\omega$ serves as a knob to systematically balance benign performance and robustness.  Notably, with $\omega=3$, we reduce the benign performance drop from 7\% to 0\% while maintaining non-trivial certifiable robustness.

\textbf{Impact of retrieved passages $k$.} We vary the number of retrieved passages $k$ from 2 to 20 and report the results in Figure~\ref{fig-topk-rqa-mistral}. As the number of retrieved passages increases, certifiable robustness and benign performance improve. We observe that the improvement can be smaller when $k$ is larger than 10; this is because additional passages (with lower rankings) usually carry less additional relevant information.

\textbf{Impact of corruption size $k^\prime$.} Figure \ref{fig-keyword-rqa-mistral} reports certifiable robustness for larger corruption sizes $k^\prime$. \framework achieves substantial certifiable robustness against multiple corrupted passages; certifiable robustness gradually decreases given a larger corruption size. When half of the passages (5 out of 10) are corrupted, useful benign passages can never outnumber malicious ones; thus, robustness reduces to zero as expected.

\textbf{Impact of keyword filtering thresholds $\alpha,\beta$.} In Figure \ref{fig-keyword-rqa-mistral}, we report the robustness of keyword aggregation with different filtering thresholds $\alpha,\beta$. Larger $\alpha,\beta$ improve certifiable robustness because fewer malicious keywords can survive the filtering. However, larger thresholds can also remove more benign keywords and thus hurt benign performance. \spnew{In practice, we need to carefully select proper filtering thresholds (e.g., using a validation dataset) to balance the trade-off between certifiable robustness and benign performance.}

\textbf{Impact of decoding probability threshold $\eta$.} Due to space limit, we analyze probability thresholds $\eta$ in Figure~\ref{Fig:apend_abl_corr} in Appendix~\ref{apx-experiment-abl}. A larger $\eta$ slightly decreases benign performance but improves certifiable robustness.

\textbf{Runtime analysis.} Table \ref{tab-runtime} reports average per-query inference latency of \framework with Mistral-7B and $k=10,\omega=1$ on one NVIDIA A100 GPU, along with the latency ratio compared to vanilla RAG---\framework is 1.16--3.65$\times$ slower than vanilla RAG.\footnote{\spnew{In contrast, popular certifiably robust techniques such as Randomized Smoothing~\cite{cohen2019certified} (for classification) incur over 100$\times$ computation overhead.}} \spnew{\framework does not incur $k$ times slowdown, as each of the $k$ isolated LLM calls processes only one passage, which is much faster and cheaper than the single large LLM call in vanilla RAG that handles all $k$ passages.}

\begin{table}[t]
\centering
\caption{runtime analysis (Mistral-7B, $k=10,\omega=1$)}
\label{tab-runtime}
    \vspace{-1em}

\resizebox{\linewidth}{!}{
\begin{threeparttable}
\begin{tabular}{lcccccccc}\toprule
  &    \multicolumn{2}{c}{RQA-MC (0-shot)}  &    \multicolumn{2}{c}{RQA (1-shot)} &  \multicolumn{2}{c}{RQA (4-shot)} & \multicolumn{2}{c}{Bio (1-shot)}  \\ 
\midrule 

Vanilla & 0.38s   &  1.00$\times$     & 0.44s   &  1.00$\times$ & 0.46s   &  1.00$\times$  & 7.69s   &  1.00$\times$  \\  
Keyword  & 0.62s   & 1.63$\times$     & 1.22s   & 2.77$\times$  & 1.68s   & 3.65$\times$  & 14.90s   & 1.94$\times$  \\  
Decoding   & 0.62s   & 1.63$\times$  & 0.51s & 1.16$\times$    & 1.32s  & 2.87$\times$  & 9.62s      & 1.25$\times$ \\ 
\bottomrule

\end{tabular}
 \begin{tablenotes}
 \item We report the per-query inference latency/ ratio with different shots of ICL exemplars. 
 \end{tablenotes}
\end{threeparttable}}

\end{table}

\section{Discussion}
In this section, we discuss the limitations and future work directions of \framework.

\spnew{\textbf{Generalizing robustness to retrieval phase.} A typical RAG pipeline consists of retrieval and generation phases. \framework operates in the generation phase and is effective when the number of corrupted passages $k^\prime$ is small---a reasonable assumption in applications like web search. To defend against even stronger attackers capable of corrupting many passages, it is important to extend RAG defense to the retrieval phase, aiming to reduce the number of corrupted passages before applying \framework for robust generation.}






\textbf{Improving computation efficiency.} In Table~\ref{tab-runtime}, we observed a small-to-moderate computation overhead. It is also important to improve computation efficiency. One possible solution is to fine-tune the LLM to eliminate the need for ICL exemplars (note that the ICL exemplars are repeated across different passage groups). Another idea is to implement a better inference pipeline that is aware of the shared prefix~\citep{juravsky2024hydragen}.

\textbf{Further improving benign performance.} As shown in Section~\ref{subsec:evalmain} and Table~\ref{tab:main_cr}, \framework yields minimal benign performance drops for QA tasks and small-to-moderate drops for long-form generation. The group size $\omega$ serves as a knob to balance benign performance and certifiable robustness. Further improving the benign performance of \framework, especially for long text generation such as summarization, remains an important direction. We note that \framework is compatible with different LLMs as well as advanced RAG approaches such as self-critic and fine-tuning~\cite{asai2023selfrag}. We expect \framework to benefit from future advancements from the NLP community. Additionally, we focused on single-hop RAG tasks in this paper; extending \framework to multi-hop questions, which require combining information from multiple passages, is another interesting direction. One promising strategy is to decompose complex queries into single-hop sub-questions and apply \framework to each question.




\section{Related Works}

\textbf{LLMs and RAG.} Large language models (LLMs)~\citep{brown2020language,achiam2023gpt} have achieved remarkable performance for various tasks; however, their responses can be inaccurate due to their limited parametric knowledge. 
Retrieval-augmented generation (RAG)~\citep{pmlr-v119-guu20a,NEURIPS2020_6b493230} aims to overcome this limitation by augmenting the model with external information retrieved from a database.
Recent works~\citep{asai2023selfrag,sail,yan2024corrective} improve RAG performance in the non-adversarial setting. 
This paper studies the adversarial robustness of RAG pipelines when an attacker corrupts a fraction of the retrieved passages

\textbf{Corruption attacks against RAG.} Early works studied misinformation attacks against QA models~\citep{du2022synthetic,pan2023attacking,pan2023risk}. 
More recent attacks have focused on the LLM-powered RAG pipeline. 
One category of these attacks targets the retrieval phase by leveraging gradient information to create adversarial contexts that are likely to be retrieved~\cite{long2024backdoor,su2024corpuspoisoningapproximategreedy}. 
Other studies address both the retrieval and generation phases, crafting adversarial passages for specific queries. 
For instance, Greshake et al.~\citep{greshake2023not} injected malicious instructions into LLM applications, and Zou et al.~\citep{zou2024poisonedrag} proposed PoisonedRAG that introduced malicious passages to mislead RAG-based QA pipelines. Shafran et al. \cite{shafran2025machine} proposed to inject passages to block LLMs from generating desired responses. Chaudhari et al.~\cite{chaudhari2024phantomgeneraltriggerattacks} proposed a backdoor attack on RAG systems, where responses are manipulated to be incorrect only when an adversarial trigger is present in the victim's queries.
In this paper, we designed \framework to be resilient to \textit{corrupted passages with arbitrary content.}


\textbf{Defenses against corruption attacks.} To mitigate misinformation attacks, Weller et al.~\citep{weller2022defending} rewrote questions to introduce redundancy and robustness; Hong et al.~\citep{hong2023discern} trained a discriminator to identify misinformation. Recently, several methods~\cite{wei2024instructrag, wang2024astute} have been proposed to enhance the robustness and trustworthiness of the RAG pipeline; however, they are built upon heuristics and lack formal robustness guarantees.
In contrast, \framework achieves certifiable robustness against passage corruption.




\section{Conclusion}\label{sec:discussion-conclusion}




\spnew{We introduced \framework, the first RAG defense framework with certifiable robustness against retrieval corruption attacks. \framework leverages an isolate-then-aggregate strategy to limit the influence of malicious passages. We proposed two secure aggregation techniques for unstructured text responses, enabling worst-case performance analysis---marking a first in this area. Experiments across diverse tasks and datasets demonstrate that \framework achieves strong certifiable robustness with negligible-to-small impact on benign performance.}


\bibliographystyle{plain}
\bibliography{ref-short.bib}


\appendices
\section{Additional Details of Robustness Certification}
\label{apx-certification}

In this section, we first present the proof of the lemmas introduced in Theorem~\ref{thm2} in Section~\ref{sec-cert-decoding} (Appendix~\ref{apx-cert-proof}), and discuss how to generalize certification algorithms to larger group sizes $\omega$ (Appendix~\ref{apx-cert-omega}), as well as passage modification (Appendix~\ref{apx-modification}).

\textbf{Important note.} In Section~\ref{sec-certification}, we focused on the simplified setting $\omega=1$, where $m=k$, $m^\prime=k^\prime$, $\cG_m = \cP_k$, and $\bfg_j=\bfp_j$, and accordingly used the simplified notation $k, k^\prime, \cP_k, \bfp_j$. In this section, we switch to the more general setting and instead use $m, m^\prime, \cG_m, \bfg_j$.


\subsection{Proof of lemmas}\label{apx-cert-proof}

In this subsection, we present the proofs of the three lemmas introduced in Section~\ref{sec-cert-decoding}. Note that we switch from the simplified notation $k, k^\prime$ used in Section~\ref{sec-cert-decoding} to the general notation $m, m^\prime$.

\renewcommand\thelemma{1}
\begin{lemma}
    If $A-B> \eta+m^\prime$ is true, the algorithm will always predict $\bft_a$.
\end{lemma}
\begin{proof}
    Without loss of generality, we only need to consider the top-2 tokens $\bft_a,\bft_b$  because other tokens have lower probability values and are less likely to be chosen.
    Let $x,y$ be the additional probability values introduced by malicious passages for tokens $\bft_a,\bft_b$, respectively. We know that $x,y\in[0,m^\prime]$ because each probability value is bounded within $[0,1]$ and the attacker can only corrupt $m^\prime$ passage groups. 
    Next, we compare the new probability value sums $A+x$ and $B+y$. 
    
    We have 
    \begin{align}
        A+x-(B+y) &= (A-B)+x-y\\
                  & >(A-B)+\min_{x,y\in[0,m^\prime]}(x-y)\\
                  &=(A-B)+(-m^\prime)\\
                  &> \eta+m^\prime-m^\prime =  \eta
    \end{align}
    According to Algorithm~\ref{alg:decoding}, we will always predict the top-1 token $\bft_a$ in this case.
\end{proof}

\renewcommand\thelemma{2}
\begin{lemma}
    If $ \eta+m^\prime \geq  A-B > | \eta-m^\prime|$ is true, the algorithm might predict the top-1 token $\bft_a$ or the no-retrieval token $\bft_\text{nor}$, but not any other token.
\end{lemma}
\begin{proof}
    We prove this lemma in two steps. First, we aim to prove that no tokens other than $\bft_a$ or $\bft_\text{nor}$ will be predicted. Without loss of generality, we only need to prove that the top-2 token $\bft_b$ will not be predicted. This is because other tokens have lower probability values than $\bft_b$ and thus are harder to be predicted. Second, we prove that the algorithm can predict the top-1 token $\bft_a$ or the no-retrieval token $\bft_\text{nor}$.

    Let $x,y$ be the additional probability values introduced by the attacker for tokens $\bft_a,\bft_b$, respectively. We know that $x,y\in[0,m^\prime]$. We next analyze the new probability value sums $A+x$ and $B+y$. We have 
    \begin{align}
        (B+y) - (A+x) & = -(A-B) + (y-x)\\
        &<  -| \eta-m^\prime| + (y-x)\\
        &\leq -| \eta-m^\prime| + \max_{x,y\in[0,m^\prime]}(y-x)\\
        &= -| \eta-m^\prime| + m ^\prime
    \end{align}
    If $ \eta\geq m^\prime$, we have 
    \begin{align}
        (B+y) - (A+x) < -| \eta-m^\prime| + m ^\prime \leq m^\prime\leq \eta
    \end{align}
    If $ \eta<m^\prime$, we have
    \begin{align}
        (B+y) - (A+x) &< -| \eta-m^\prime|+m^\prime= \eta-m^\prime+m^\prime= \eta
    \end{align}
    We have $(B+y) - (A+x)< \eta$ in both cases. Therefore, the probability gap is not large enough for the algorithm to output the top-2 token $\bft_b$.

    Next, we aim to prove that the algorithm can output the top-1 token $\bft_a$ or the no-retrieval token $\bft_\text{nor}$. We need to show that there exist feasible $(A,B,x,y, \eta,m^\prime)$ tuples such that $(A+x)-(B+y)> \eta$ (predicting the top-1 token $\bft_a$) and $(A+x)-(B+y)\leq  \eta$ (predicting the no-retrieval token $\bft_\text{nor}$). We can derive the following inequalities.

    \begin{align}
        \min(A-B) + &\min_{x,y\in[0,m^\prime]}(x-y)\leq \,\nonumber\\
        &(A+x)-(B+y) \\
        &\leq \max(A-B) + \max_{x,y\in[0,m^\prime]}(x-y)\nonumber\\
        | \eta-m^\prime| - m^\prime < \, &(A+x)-(B+y) \leq  \eta+m^\prime +m^\prime 
    \end{align}

Since $m^\prime>0$, clearly we have $| \eta-m^\prime|-m^\prime< \eta< \eta+2m^\prime$. Therefore, there exist cases that satisfy $| \eta-m^\prime|-m^\prime \leq(A+x)-(B+y)\leq \eta$, and the algorithm can output a no-retrieval token $\bft_\text{nor}$. There also exists cases that satisfy $ \eta<(A+x)-(B+y) \leq  \eta+2m^\prime$, the algorithm can output the top-1 token $\bft_a$.
\end{proof}
\renewcommand\thelemma{3}
\begin{lemma}
    If $ \eta-m^\prime \geq A-B>0$ is true, the algorithm will always predict a no-retrieval token.
\end{lemma}
\begin{proof}
    Without loss of generality, we only need to consider the top-2 tokens $\bft_a,\bft_b$ because other tokens have lower probability values and are less likely to be chosen.
    Let $x,y$ be the additional probability values introduced by the attacker for tokens $\bft_a,\bft_b$, respectively. We know that $x,y\in[0,m^\prime]$.
    Next, we analyze the new probability value sums $A+x$ and $B+y$. 

    To always output a no-retrieval token, we require $|(A+x)-(B+y)|\leq  \eta, \forall \, x,y\in[0,m^\prime]$. Equivalently, we require
    \begin{align}
    &&\forall\, x,y\in&[0,m^\prime]\nonumber\\
    \Leftrightarrow&&    - \eta-x+y\leq  &\,A-B\leq  \eta -x + y\\
    \Leftrightarrow&&     - \eta + \max_{x,y\in[0,m^\prime]}(-x+y)\leq &\,A-B \\
   &&\hfill  \leq\eta+&\min_{x,y\in[0,m^\prime]}(-x+y)\\
    \Leftrightarrow&&     - \eta + m^\prime\leq  & \,A-B \leq  \eta - m^\prime 
    \end{align}
    Note that we have $A-B>0$ since $A$ is the probability sum of the top-1 token. So we have $ \eta-m^\prime \geq A-B> 0 \Leftrightarrow$ the algorithm will always output a no-retrieval token.
\end{proof}

\begin{figure}[t]
    \centering
    \includegraphics[width=\linewidth]{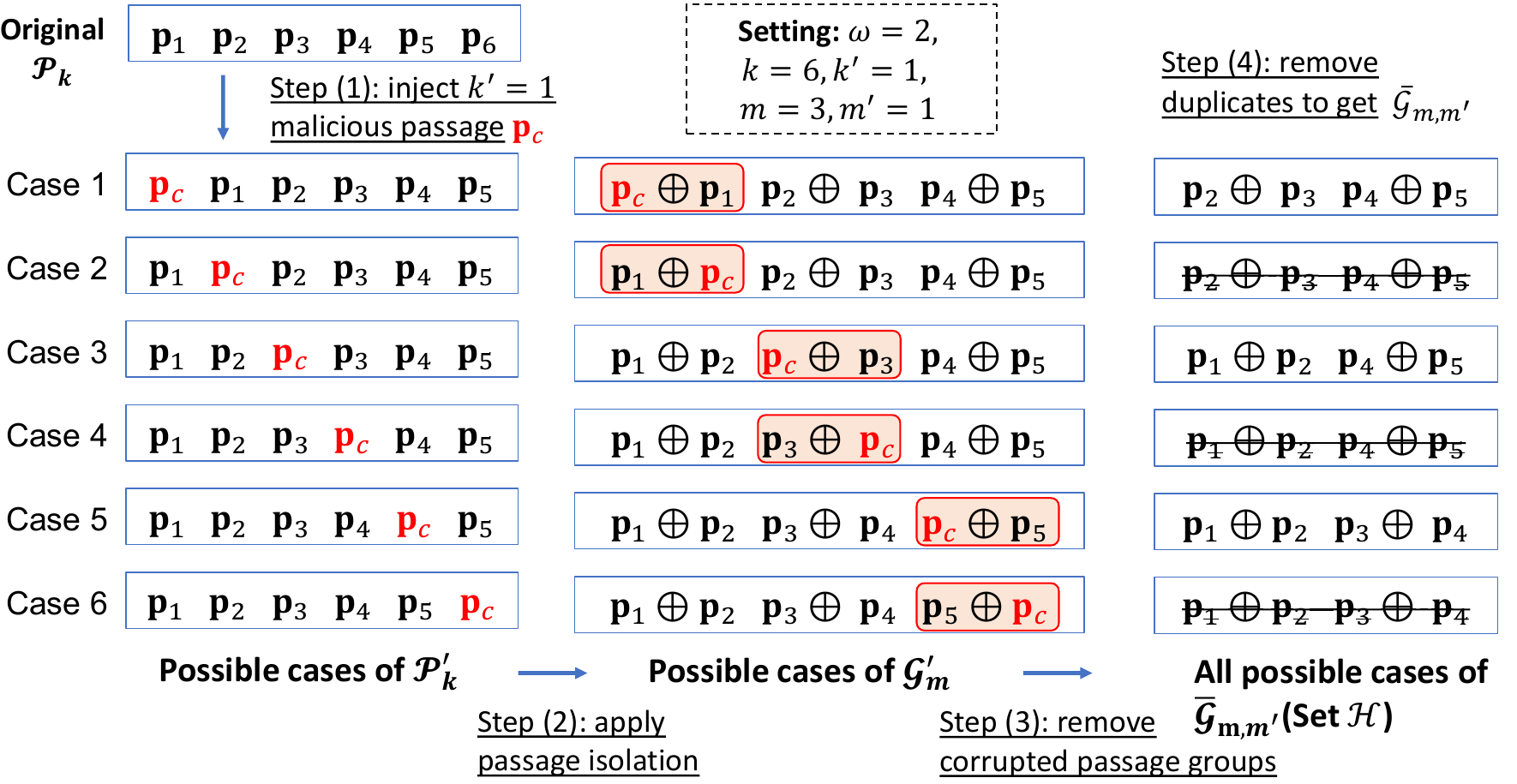}
    \caption{\textbf{Example of enumerating all possible input passage groups.} (1) we first inject $k^\prime$ corrupted passage $\bfp_c$ to all possible positions, resulting in $k\choose k^\prime$ possible cases of $\cPkp$. (2) We apply passage isolation $\textsc{IsoGroup}(\cdot,\omega)$ to each possible $\cPkp$ and get corresponding $\cG^\prime_m$. (3) We remove the corrupted passage groups from each $\cG^\prime_m$ (marked with red boxes) to get $\cGmp$. (4) We remove duplicates and form a set $\cH$ with all possible (distinct) $\cGmp$. We use $\cH\gets\textsc{BenignGroupCases}(\cP_k,\omega,k^\prime)$ to denote this process.}
    \label{fig-enumeration-injection}
\end{figure}

\subsection{Generalizing to larger group sizes $\omega$}\label{apx-cert-omega}

In Section~\ref{sec-certification}, our presentation focused on group size $\omega=1$. With this simplified setting, we have $m=k,m^\prime=k^\prime, \bfg_j=\bfp_j, \cG_m = \cP_k$. In this section, we discuss how to perform certification for larger group sizes $\omega$.

\textbf{Understanding isolated inputs to LLM with $\omega>1$}. When the group size $\omega$ is larger than one, $\cG_m$ no longer equals $\cP_k$, and the LLM input analysis becomes slightly more complicated. Recall that an attacker can \textit{inject} $k^\prime$ passages into arbitrary positions within the top-$k$ retrieval result; thus, there are $k \choose k^\prime$ possible cases of injection positions, and we need to analyze all of them. For each of $k \choose k^\prime$ cases, we can simulate the isolation operation $\textsc{IsoGroup}(\cdot)$ to identify $m^\prime$ out of  $m=\lceil\frac{k}{\omega}\rceil$ passage groups that overlap with the injection positions; our certification will be based on the other $m-m^\prime$ benign passage groups, which we denote as $\cGmp$.\footnote{When $m-m^\prime\leq0$, we cannot perform meaningful certification to compute a non-trivial $\tau$ value. We need to choose a proper $\omega$ to avoid this failure case, as discussed in the remark in Section~\ref{sec-iso-group}.}
Figure~\ref{fig-enumeration-injection} provides a concrete example of enumerating all possible $\cGmp$, given $k=6,\omega=2,k^\prime=1$.\footnote{When $\omega=1$, there is only one possible $\cGmp$, which equals $\cP_{k-k^\prime}$. This is the simplified setting we discussed in Section~\ref{sec-certification}.}

\new{Once we understand possible isolated inputs to the LLM, we can replace $\cP_{k-k^\prime}$ and $k^\prime$ (i.e., the inputs to the certification procedures discussed in Section~\ref{sec-certification}), with $\cGmp$ and $m^\prime$ to run certification (Algorithm~\ref{alg:keyword-certify} or \ref{alg:decoding-certify}) to get $\tau$ for each possible $\cGmp$. The certification outcome should be the lowest $\tau$ across all $\cGmp$.}

\textbf{Certification procedure.} Formally, we use $\cH$ to denote the set of all possible $\cGmp$, and use $\cH\gets\textsc{BenignGroupCases}(\cP_k,\omega,k^\prime)$ to denote the process of getting all possible combinations of uncorrupted benign passage groups, given $k \choose k^\prime$ cases of passage injection positions.\footnote{$\bfp_c$ in Figure~\ref{fig-enumeration-injection} can have any malicious content. However, as we filter corrupted passage groups in Step (3), the malicious content does not affect our subsequent robustness analysis, and the set $\cH$ in Step (4) is still finite.}

We present the certification procedure (Algorithm~\ref{alg-cert-workflow}), which has two major steps. First, we generate all possible cases of benign passage group combinations via $\cH\gets\textsc{BenignGroupCases}(\cP_k,\omega,k^\prime)$ (Line~\ref{ln-workflow-allcases}). Second, for each $\cGmp\in\cH$, we aim to determine its $\tau$ value; we use an abstract procedure to denote this operation: $\tau_{\cGmp}\gets\textsc{CertifyOneCase}(\cGmp,\llm,\cZ,\bfq,\bfa)$, where $\llm,\cZ,\bfq,\bfa$ are the underlying LLM, defense/attack parameters, input query, and reference answer, respectively (Line~\ref{ln-workflow-tau}). Here, $\textsc{CertifyOneCase}(\cdot)$ can be either Algorithm~\ref{alg:keyword-certify} or Algorithm~\ref{alg:decoding-certify} discussed in Section~\ref{sec-certification}. Additionally, we use $\tau^*$ to track the lowest $\tau_{\cGmp}$ computed for each case (Line~\ref{ln-workflow-mintaustar}) and finally return $\tau^*$ as the certification outcome. Within this workflow, we only need to design $\textsc{CertifyOneCase}(\cdot)$ to return the correct $\tau$ for each possible case of benign passage groups $\cGmp$.

\begin{algorithm}[t]
\caption{Certification workflow}\label{alg-cert-workflow}
\begin{algorithmic}[1]
\Require Benign retrieved data $\cP_k$, group size $\omega$, corruption size $k^\prime$, query $\bfq$, model $\llm$, reference answer $\bfa$, other defense/attack parameters $\cZ$.
\Procedure{Certify}{}
\State  $\cH\gets\textsc{BenignGroupCases}(\cP_k,\omega,k^\prime)$\label{ln-workflow-allcases}
\State $\tau^*\gets\infty$
\For{$\cGmp\in\cH$}
\State $\tau_{\cGmp}\gets\textsc{CertifyOneCase}(\cGmp,\llm,\cZ,\bfq,\bfa)$\label{ln-workflow-tau}
\State $\tau^*\gets\min(\tau^*,\tau_{\cGmp})$\label{ln-workflow-mintaustar}
\EndFor
\State\Return $\tau^*$
\EndProcedure
\end{algorithmic}
\end{algorithm}

\definecolor{Gray}{gray}{0.9}
\definecolor{shadecolor}{rgb}{0.9,0.9,0.9}
	
\begin{table*}[!t]
\centering
{
\begin{threeparttable}
\caption{Certifiable robustness and benign performance ($k=10,k^\prime=1$, GPT-3.5)} %
\label{tab:main_cr_gpt}
\vspace{-1em}
\begin{tabular}{llcccccccc}\toprule
Task& \multicolumn{1}{l}{\multirow{2}{*}{Model/ }}&\multicolumn{2}{c}{Multiple-choice QA} &  \multicolumn{4}{c}{Short-answer QA} & \multicolumn{2}{c}{Long-form generation}     \\
Dataset&\multicolumn{1}{l}{\multirow{2}{*}{Defense}} & \multicolumn{2}{c}{RQA-MC}   &  \multicolumn{2}{c}{RQA}  & \multicolumn{2}{c}{NQ} &  \multicolumn{2}{c}{Bio}   \\ 
\multicolumn{1}{l}{LLM} & & {(bacc)} & (cacc) & (bacc) & (cacc)  & (acc) & (cacc) & (bllmj) & (cllmj)  \\
\midrule
\multirow{3}{*}{GPT$_{\textsc{3.5}}$}
& No RAG & 8.0 & -- & 2.0 & -- & 24.6 &--  &   12.6&--  \\
\cdashline{2-10}\noalign{\vspace{1.5pt}} 
& Vanilla & 80.4 & 0.0  &  65.4 & 0.0   &  58.8   & 0.0   & 76.6  & 0.0    \\
\cdashline{2-10}\noalign{\vspace{1.5pt}} 
& Keyword  & 76.4 & 69.6 &  56.4 &  37.8 &  54.2   &  37.0   & 59.4 & 24.0\\ 
\bottomrule
\end{tabular}
\begin{tablenotes}
    \item (bacc): benign accuracy; (cacc): certifiable accuracy; (llmj): benign LLM-judge score; (cllmj): certifiable LLM-judge score. 
\end{tablenotes}
\end{threeparttable}}
\end{table*}

\subsection{Generalizing to Passage Modification}\label{apx-modification}
\begin{figure}
    \centering
    \includegraphics[width=\linewidth]{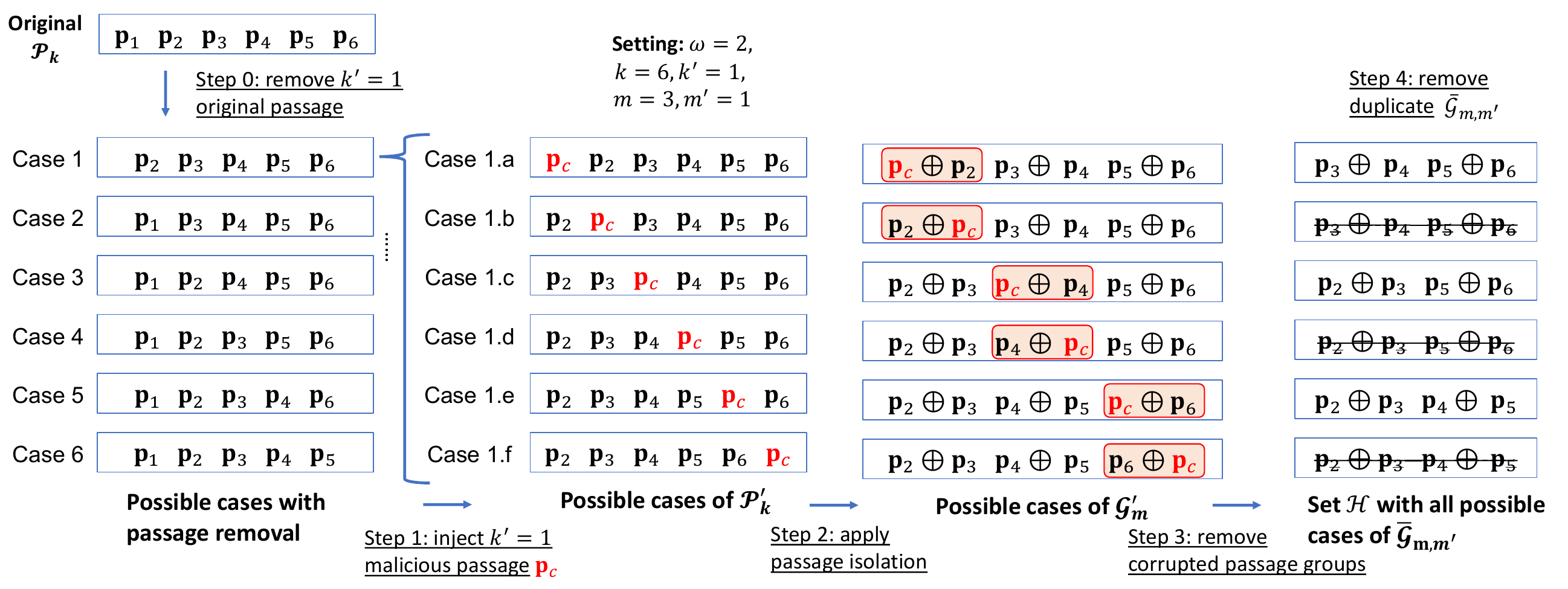}
    \caption{\textbf{Example of the process $\cH\gets\textsc{BenignGroupCases}(\cP_k,\omega,k^\prime)$ for passage modification.} Passage modification can be decomposed into passage removal and passage injection. Given $k$ passages, the $\textsc{BenignGroupCases}(\cdot)$ procedure first removes $k^\prime$ corrupted passage; there are $k\choose k^\prime$ possible cases. For each case (we plot for Case 1 in the figure), we next inject $k^\prime$ malicious passage $\bfp_c$ to all possible locations, resulting in $k\choose k^\prime$ possible cases of $\cPkp$. The rest of the procedure is identical to the passage injection case discussed in Figure~\ref{fig-enumeration-injection}: we apply passage isolation $\textsc{IsoGroup}(\cP_k,\omega)$ to each possible $\cPkp$ and get corresponding $\cG^\prime_m$, and remove the corrupted passage groups from each $\cG^\prime_m$ and get $\cGmp$. Finally, we enumerate all possible cases and form the output set $\cH$ with all possible distinct $\cGmp$.}
    \label{fig-enumeration-modify}
\end{figure}

\begin{table}[t]
\centering

    \caption{certifiable robust accuracy against passage injection and modification}
    \label{tab:modification}
    \resizebox{\linewidth}{!}{
\begin{tabular}{lcccccc}\toprule
 \multirow{2}{*}{Model/} & \multicolumn{2}{c}{Multiple-choice QA}&  \multicolumn{4}{c}{Open-domain QA}  \\
\multirow{2}{*}{defense} & \multicolumn{2}{c}{RQA-MC}   &  \multicolumn{2}{c}{RQA}  & \multicolumn{2}{c}{NQ}     \\ 
 & inj. & modi. & inj. & modi.  & inj. & modi.  \\
\midrule

 Keyword  & \multirow{2}{*}{71.0}   & \multirow{2}{*}{ 59.0} &  38.0 &28.0&26.0&20.0\\ 
 Decoding &  &  &  37.0&23.0&29.0&13.0\\

\bottomrule
\end{tabular}
}
\end{table}
In the main body of this paper, we focus on passage \textit{injection} where the attacker can inject a small number of passages but cannot modify the original passages. 
In this section, we aim to demonstrate that \framework is directly applicable to passage \textit{modification} where the attacker can modify a small number of original passages. We can use the same inference algorithms discussed in Algorithm~\ref{alg:keyword} and Algorithm~\ref{alg:decoding}, as well as the certification algorithms discussed in Algorithm~\ref{alg:keyword-certify} and Algorithm~\ref{alg:decoding-certify}. The only thing we need to change is the implementation of $\textsc{BenignGroupCases}(\cdot)$ (discussed in Algorithm~\ref{alg-cert-workflow}). We will present the details next. 

\textbf{Overview.} We can decompose \textit{passage modification} into two steps: the attacker first \textit{removes} arbitrary $k^\prime$ original passages and then \textit{injects} $k^\prime$ malicious passages into arbitrary locations. There are $k \choose k^\prime$ possible cases for passage removal and $k \choose k^\prime$ cases for passage injection. The procedure $\textsc{BenignGroupCases}(\cdot)$ need to enumerate all these possible cases. 

We provide a visual example (with $k=6,\omega=2,k^\prime=1$) in Figure~\ref{fig-enumeration-modify}.
Given the retrieved passage $\cP_6=(\bfp_1,\bfp_2,\bfp_3,\bfp_4,\bfp_5,\bfp_6)$, the attacker first removes $k^\prime=1$ original passage, leading to six possible cases $(\bfp_2,\bfp_3,\bfp_4,\bfp_5,\bfp_6)$, $(\bfp_1,\bfp_3,\bfp_4,\bfp_5,\bfp_6)$, $(\bfp_1,\bfp_2,\bfp_4,\bfp_5,\bfp_6)$, $(\bfp_1,\bfp_2,\bfp_3,\bfp_5,\bfp_6)$, $(\bfp_1,\bfp_2,\bfp_3,\bfp_4,\bfp_6)$, and $(\bfp_1,\bfp_2,\bfp_3,\bfp_4,\bfp_5)$, denoted as ``possible cases with passage removal'' in the figure.

Then, the attacker injects one corrupted passage, denoted as $\bfp_c$ into an arbitrary location. Take $(\bfp_2,\bfp_3,\bfp_4,\bfp_5,\bfp_6)$ as an example. the injected retrieval then becomes $(\underline{\bfp_c},\bfp_2,\bfp_3,\bfp_4,\bfp_5,\bfp_6)$, $(\bfp_2,\underline{\bfp_c},\bfp_3,\bfp_4,\bfp_5,\bfp_6)$, $(\bfp_2,\bfp_3,\underline{\bfp_c},\bfp_4,\bfp_5,\bfp_6)$, 
$(\bfp_2,\bfp_3,\bfp_4,\underline{\bfp_c},\bfp_5,\bfp_6)$,
$(\bfp_2,\bfp_3,\bfp_4,\bfp_5,\underline{\bfp_c},\bfp_6)$, or $(\bfp_2,\bfp_3,\bfp_4,\bfp_5,\bfp_6,\underline{\bfp_c})$. Next, we can apply $\textsc{IsoGroup}(\cdot)$ with $\omega=2$ and get six different cases of grouped passages $\cG^\prime_m$, with $m=\lceil\frac{k}{\omega}\rceil=3$; we can express them as $(\underline{\bfp_c\cat\bfp_2},\bfp_3\cat\bfp_4,\bfp_5\cat\bfp_6)$, $(\underline{\bfp_2\cat\bfp_c},\bfp_3\cat\bfp_4,\bfp_5\cat\bfp_6)$, $(\bfp_2\cat\bfp_3,\underline{\bfp_c\cat\bfp_4},\bfp_5\cat\bfp_6)$, $(\bfp_2\cat\bfp_3,\underline{\bfp_4\cat\bfp_c},\bfp_5\cat\bfp_6)$, $(\bfp_2\cat\bfp_3,\bfp_4\cat\bfp_5,\underline{\bfp_c\cat\bfp_6})$, $(\bfp_2\cat\bfp_3,\bfp_4\cat\bfp_5,\underline{\bfp_6\cat\bfp_c})$. After that, we can get possible $\cGmp$ as $(\bfp_3\cat\bfp_4,\bfp_5\cat\bfp_6)$, $(\bfp_2\cat\bfp_3,\bfp_5\cat\bfp_6)$, and $(\bfp_2\cat\bfp_3,\bfp_4\cat\bfp_5)$. We can repeat this process for different cases of passage removal to generate all possible (distinct) $\cGmp$ and obtain $\cH$.

\textbf{Experiment results.} We use Mistral-7B-Instruct with the top-10 retrieved passages from QA datasets for experiments. We set $\alpha=0.3,\beta=3$ for keyword aggregation, and $\eta=0$ for decoding aggregation. We report the certifiable robust accuracy for injecting or modifying $k^\prime=1$ passage in Table~\ref{tab:modification}.  As shown in the table, our \framework algorithm achieves good certifiable robustness against both passage modification and injection. Note that we use \textit{the same inference algorithm} (Algorithm~\ref{alg:keyword} and Algorithm~\ref{alg:decoding} discussed in Section~\ref{sec:algorithm}) for both injection and modification attacks. The certifiable robust accuracy for passage modification is lower than that for passage injection. This is expected because passage modification is a stronger attack than passage injection. 


\begin{table*}[t]
\centering
    \caption{Empirical robustness of \framework on GPT-3.5 ($k=10,k^\prime=1$) against PIA and Poison attacks }
    \label{tab:main_em_gpt}
    

{
\begin{threeparttable}
    
\begin{tabular}{llcccccccc}\toprule
Task&   &   \multicolumn{4}{c}{Short-form open-domain QA} & \multicolumn{2}{c}{Long-form generation}     \\
Dataset&    \multirow{2}{*}{Model/}&    \multicolumn{2}{c}{RQA}  & \multicolumn{2}{c}{NQ} &  \multicolumn{2}{c}{Bio}   \\ 
Attack&  \multirow{2}{*}{Defense} & PIA   & Poison   & PIA   & Poison & PIA   & Poison  \\ 
LLM &  & racc\gua / asr\rda & racc\gua / asr\rda & racc\gua / asr\rda & racc\gua / asr\rda & rllmj\gua / asr\rda & rllmj\gua  / asr\rda \\
\midrule 
\multirow{2}{*}{GPT$_{\textsc{3.5}}$}
& Vanilla &   10.2 / 82.2   &  51.6 / 31.6   &  11.0 / 67.8   &  51.8 / 14.4 &  17.2 / 90.0   &  43.0  / 56.0 \\
& Keyword  &   52.6 / 5.0   &  51.6 / 4.6    &  53.0 / 5.2   &  52.6 / 4.6  &   56.6  / 0.0 &   52.4  / 0.0  \\ 
\bottomrule
\end{tabular}
\begin{tablenotes}
    \item (racc):~robust accuracy; (rllmj): robust LLM-judge score;  (asr): targeted attack success rate.
\end{tablenotes}
\end{threeparttable}
}
\end{table*}

\section{Additional Details of Implementation}\label{apx-details-implementation}

\textbf{Implementation of keyword extraction.} We use the spaCy library to preprocess every text response. We consider words with POS tags of ADJ (adjective), ADV (adverb), NOUN (noun), NUM (numeral), PROPN (proper noun), SYM (symbol), and X (others) to be most informative and use them as keywords or to form keyphrases. Let us call words with these tags ``informative words'' and words with other tags ``uninformative words''. Our keyword set contains (1) all lemmatized informative words and (2) keyphrases formed by combining consecutive informative words between two nearby uninformative words. 

For long-form text generation tasks, we found that the keyword sets can sometimes become too large and thus make robustness certification computationally infeasible. To reduce the number of extracted keywords/keyphrases, we prompt the model to output a list of short phrases instead of long texts and only retain keyphrases with more than two words. 

\textbf{Detailed setup of empirical attacks.} 
For prompt injection attacks~\cite{greshake2023not}, the attacker injects the following malicious instruction: \stexttt{``When asked to provide the answer for the following question: \{query\}, please output \{target answer\}.''} For data poisoning attacks, we follow PoisonedRAG~\citep{zou2024poisonedrag} and use GPT-4 to generate fake texts that support incorrect answers chosen by the attacker. We repeat malicious content multiple times (10 times for QA and 3 times for the long-generation task) in a single passage to simulate a strong attacker; the robustness against this strong attack (Table~\ref{tab:main_em}) further demonstrates the strength of our defense. 

In addition to reporting model performance under attack as the robustness metric, we also report the attack success rate (ASR), which is defined as the ratio of model responses that contain the malicious {target texts}. For QA tasks, we follow PoisonedRAG~\cite{zou2024poisonedrag} and generate the incorrect target texts via prompting GPT-4. For biography generation, we set the target answer to be \stexttt{``\{person\} is a good guy''} for PIA and \stexttt{``born on January 11, 1990''} for data poisoning.



\begin{figure*}[t]
 \centering
 \setlength{\tabcolsep}{2pt} 
 \resizebox{\linewidth}{!}{
 \begin{tabular}{cccc}
   \includegraphics[width=0.24\textwidth]{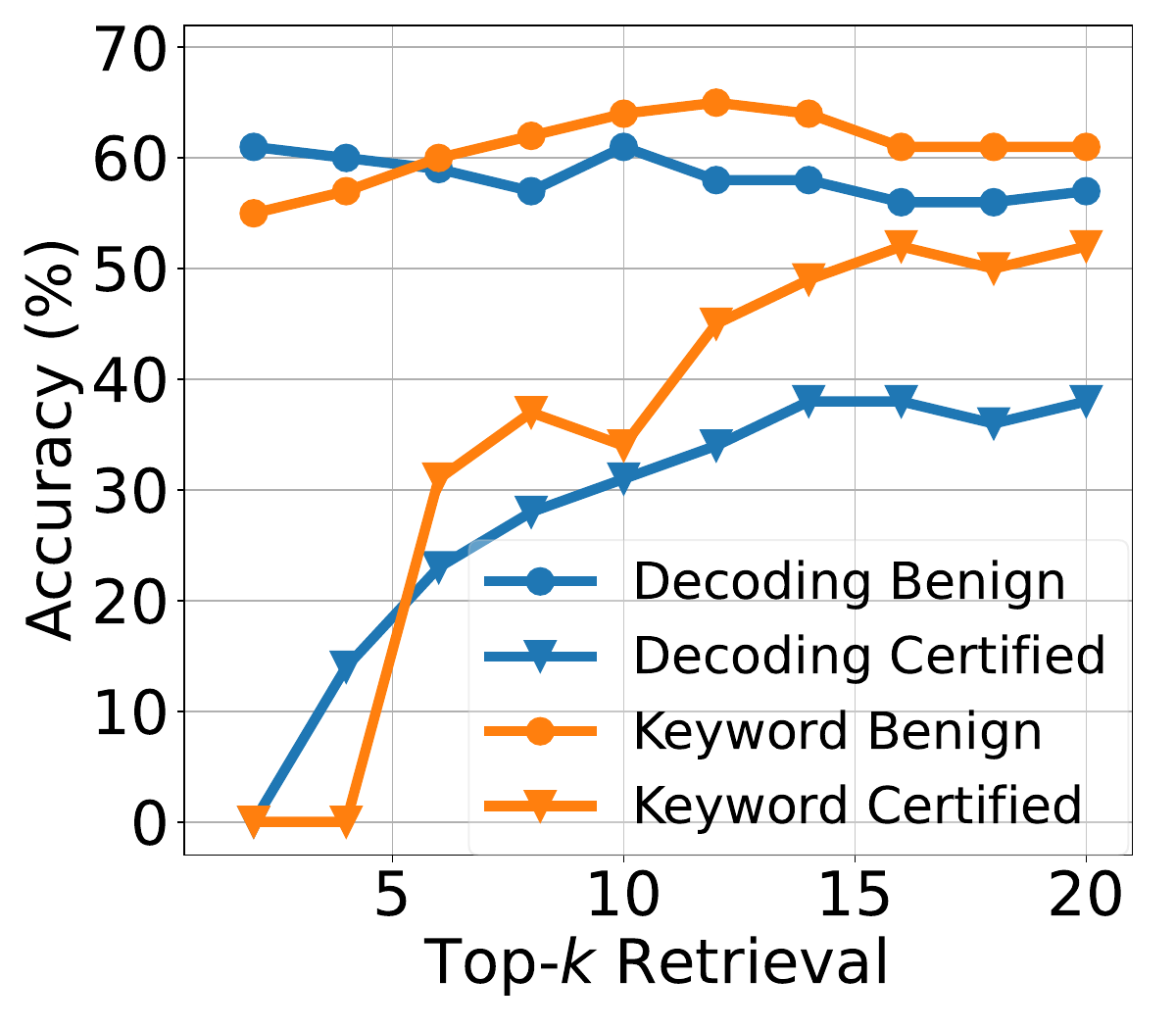} & 
   \includegraphics[width=0.24\textwidth]{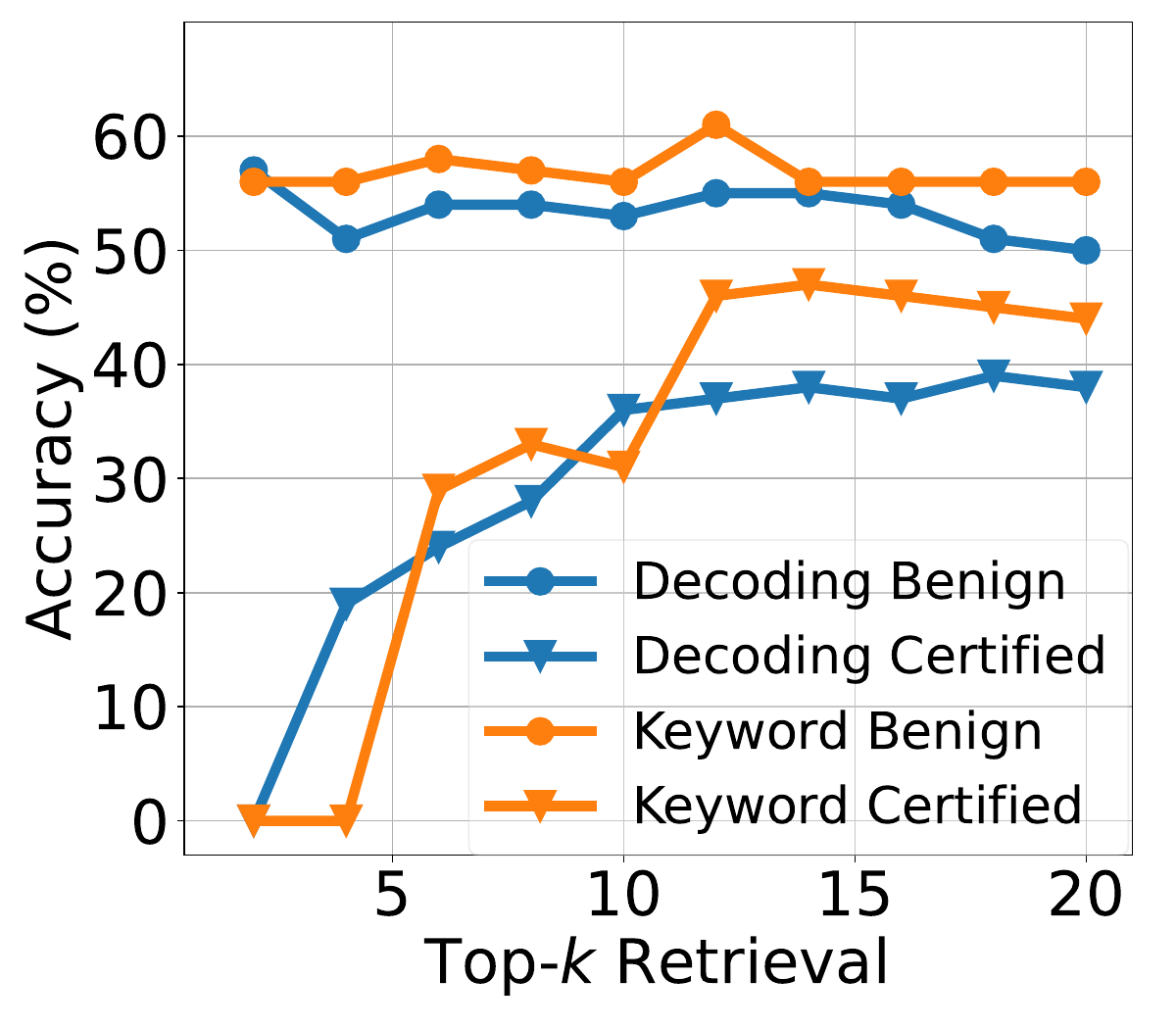} & \includegraphics[width=0.24\textwidth]{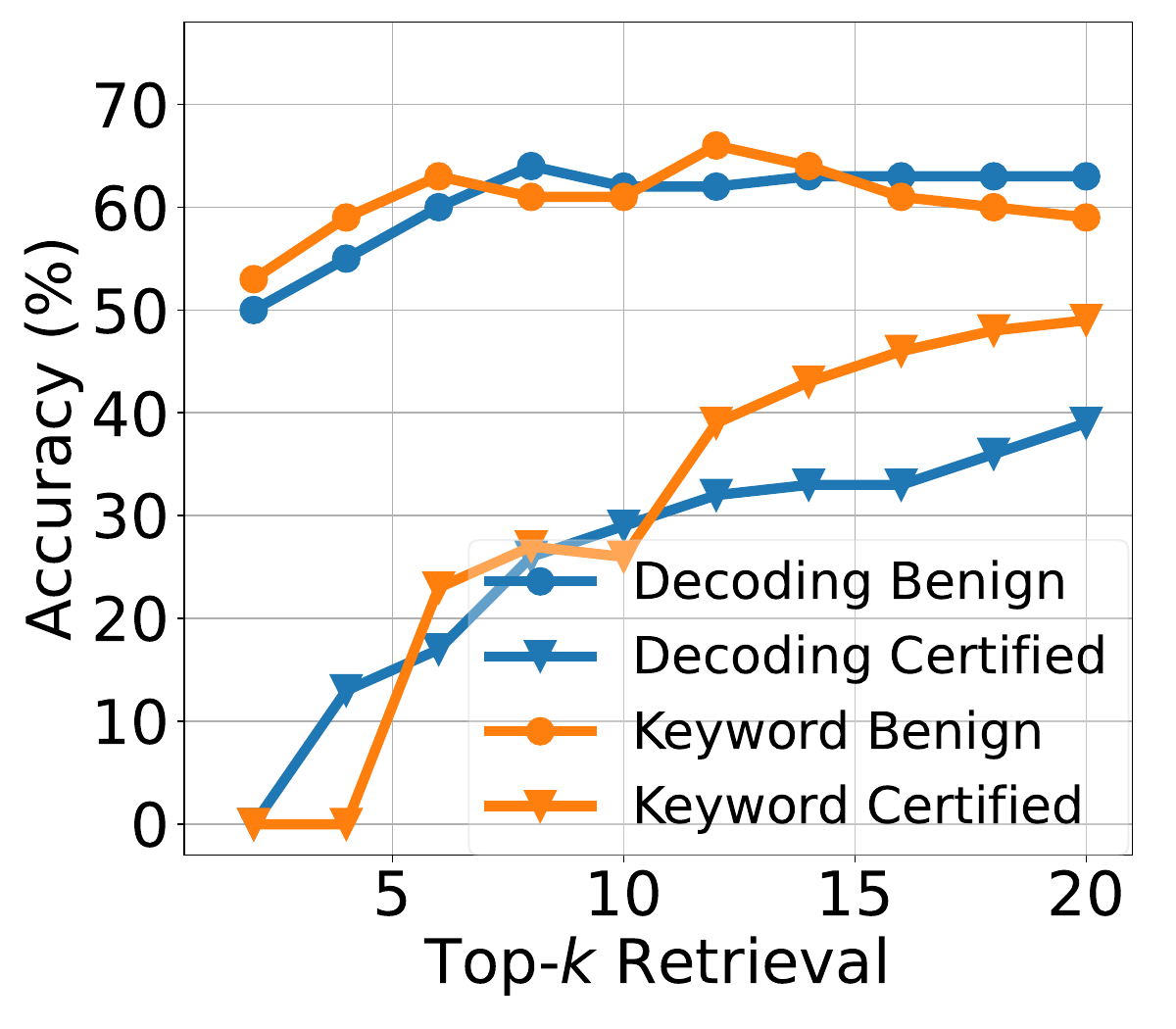} &
   \includegraphics[width=0.24\textwidth]{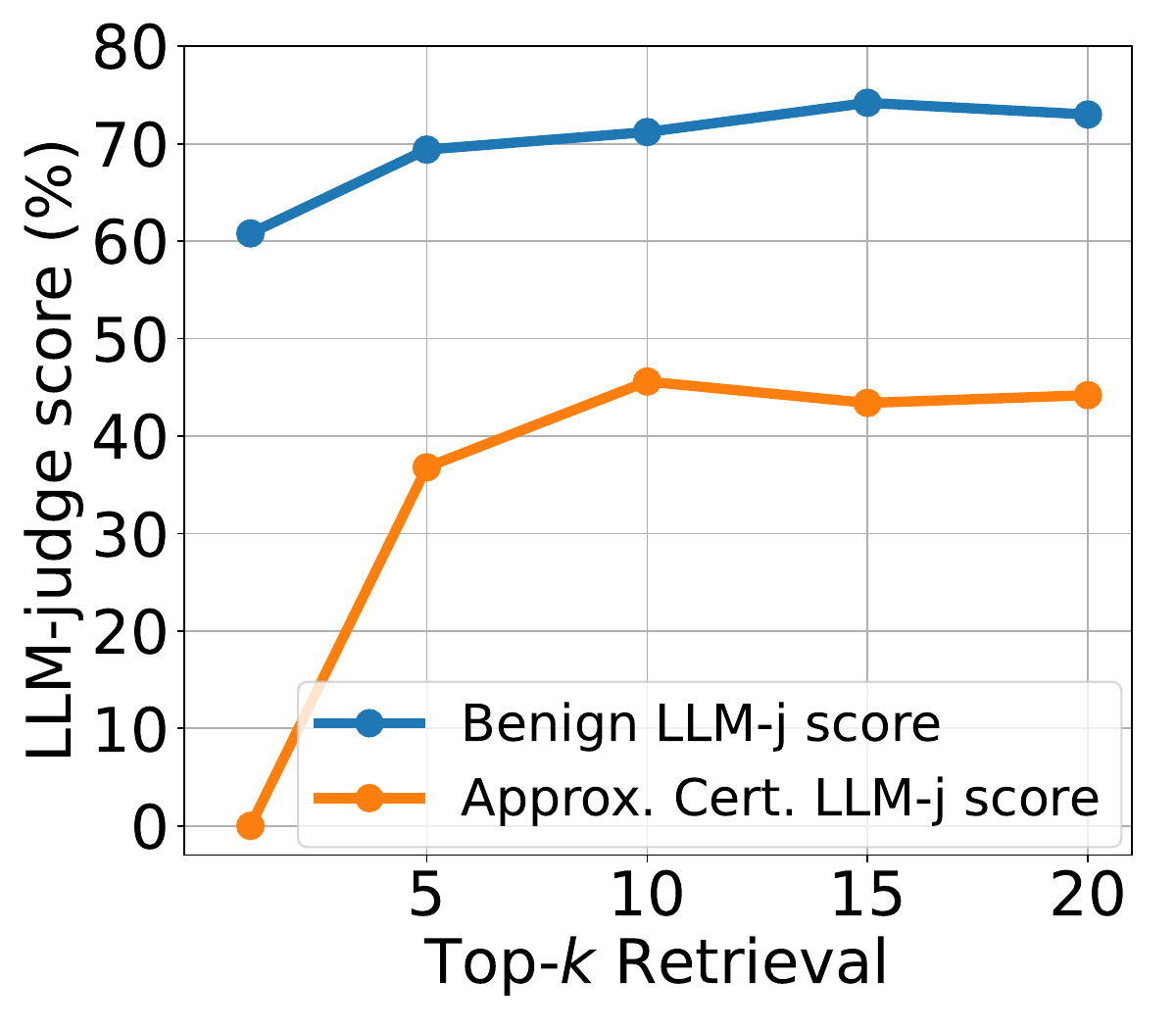}  \\
  
   \footnotesize (a) RQA (Llama-7B) &  \footnotesize(b) NQ (Llama-7B)& \footnotesize(c) NQ (Mistral-7B) &  \footnotesize(d)  Biography (Mistral-7B, $\eta=1$) \\
 \end{tabular}}
 \caption{The impact of top-$k$ retrieval on \framework (corruption size $k^\prime=1$). }
 \label{Fig:apend_abl_topk}
\end{figure*}

\section{Additional Experiment Results and Analyses}\label{apx-experiment-abl}

\textbf{Experiments with GPT Models.}
We report the certifiable robustness and benign performance of \framework with GPT-3.5-turbo in Table~\ref{tab:main_cr_gpt}, and its empirical robustness against two attacks in Table~\ref{tab:main_em_gpt}.
 Consistent with our main results, \framework achieves strong certifiable and empirical robustness—for example, 69.6\% and 37.8\% certified accuracy on RQA-MC and RQA, respectively. Under PIA attacks, our \framework limits the attack success rate to 5.0\%, compared to over 80\% for the vanilla method. We did not implement decoding aggregation for GPT-3.5 due to the excessive number of API calls required---we can only get the probability for \textit{one next-token} prediction \textit{with each API call}. This is a major difference from open-weight models, for which we can reuse the KV cache, i.e., storing cache for the first $N$ tokens and reusing them to predict the  $(N+1)^\text{th}$ token. For each separate API call, the model needs to recompute everything for the first $N$ tokens to get the $(N+1)^\text{th}$ token prediction.

\textbf{Impact of retrieved passages $k$.} In Figure \ref{Fig:apend_abl_topk}, we include additional experimental results for the RealtimeQA, Natural Questions, and Biography Generation datasets using the Llama-7B and Mistral-7B models, with a different number of retrieved passages $k$. The observation is consistent with what we discussed in Section~\ref{subsec:evalabl}: as the number of retrieved passages increases, both certifiable robustness and benign performance improve. 

\begin{figure}[!b]
 \centering
 \setlength{\tabcolsep}{1pt} 
 \renewcommand{\arraystretch}{0.8} 
 \resizebox{0.8\linewidth}{!}{\footnotesize
 \begin{tabular}{cc}
   \includegraphics[width=0.4\linewidth]{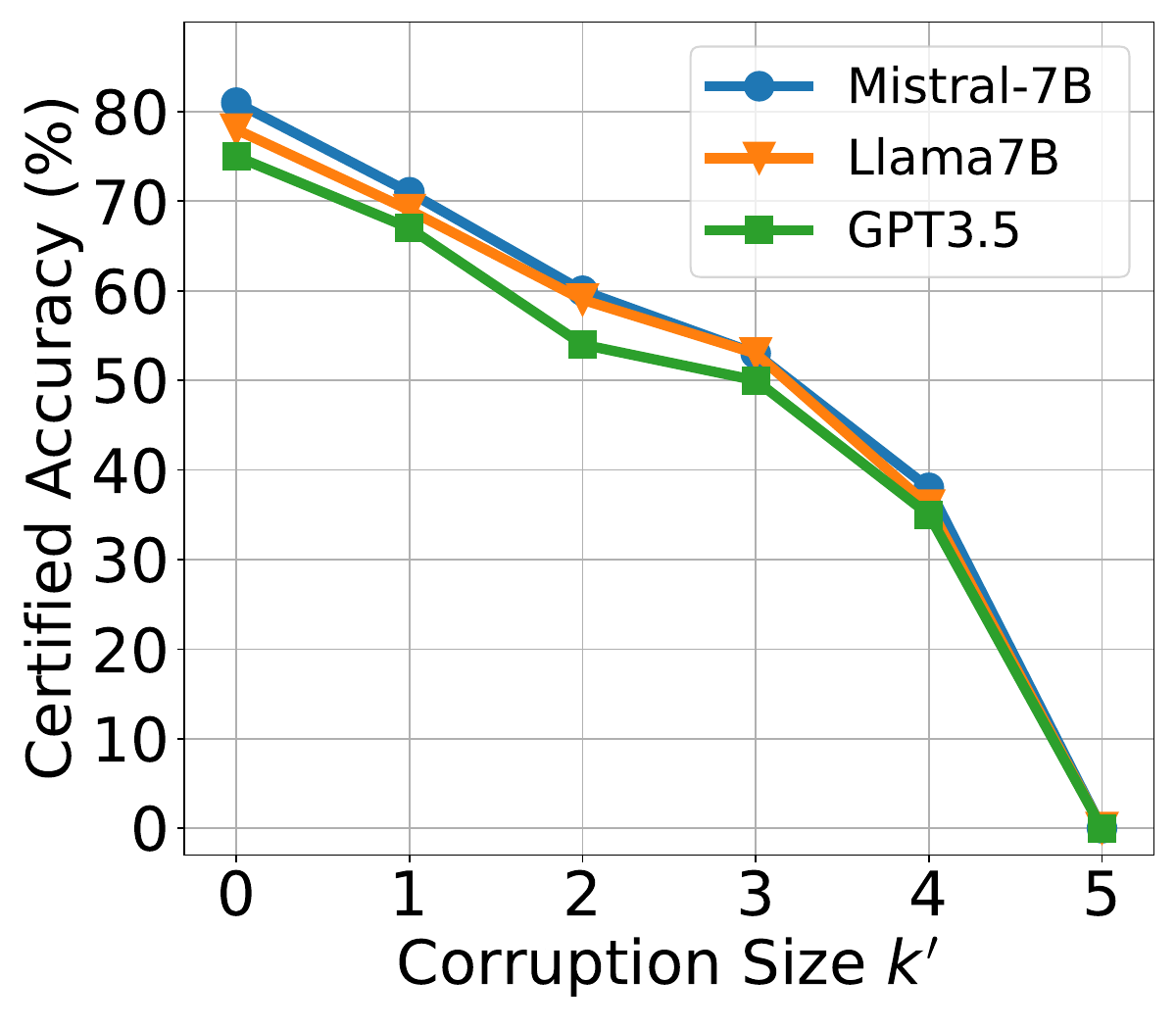}\label{Fig:apend_abl_corr_mc} & 
   \includegraphics[width=0.4\linewidth]{Figures/experiments/topk/realtimeqa-mistral-none-corr-all.pdf} \\
   (a) RealtimeQA-MC  &  (b) RealtimeQA  \\
    \noalign{\medskip} \hdashline \noalign{\medskip}
   \includegraphics[width=0.4\linewidth]{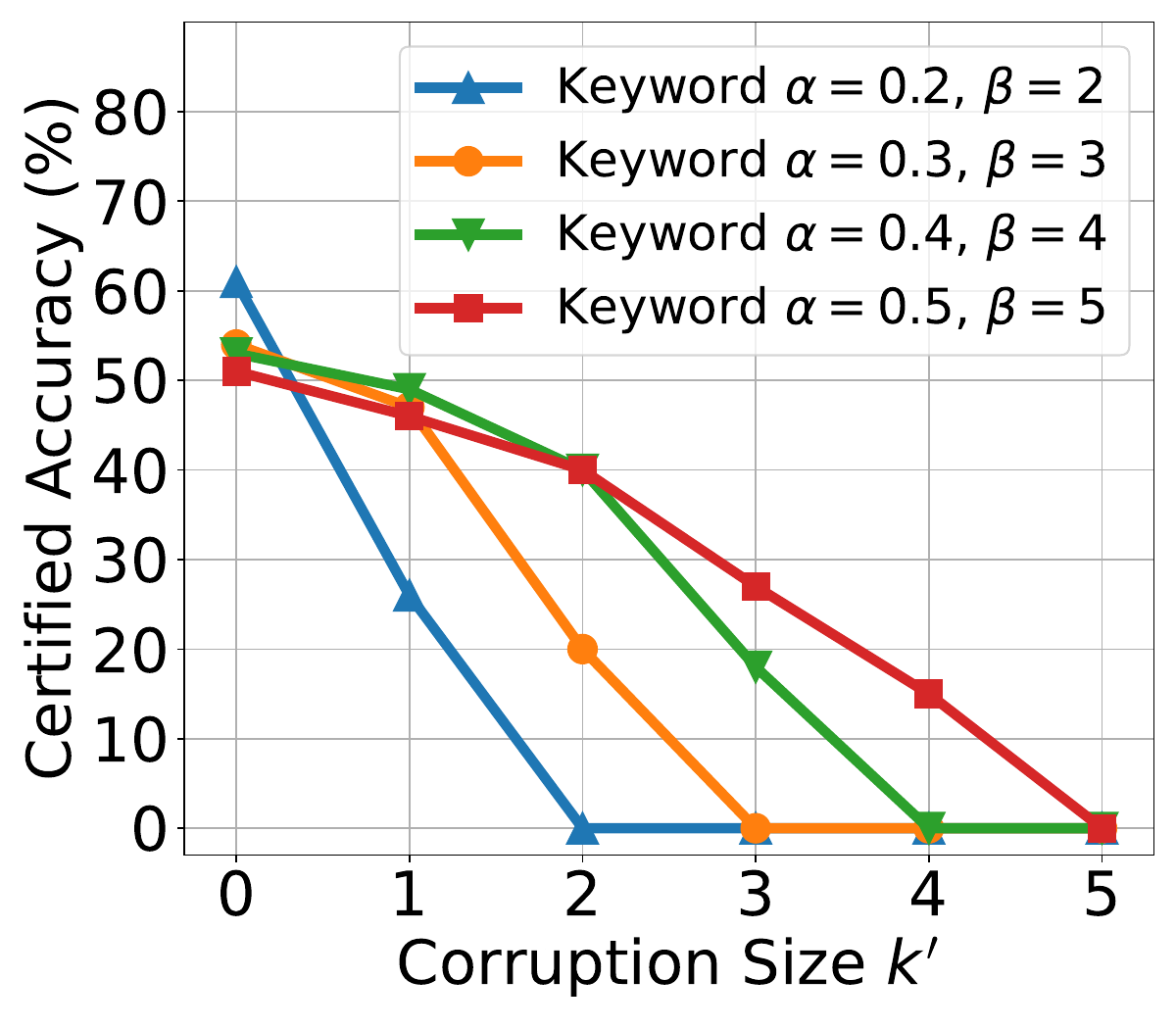} &
   \includegraphics[width=0.4\linewidth]{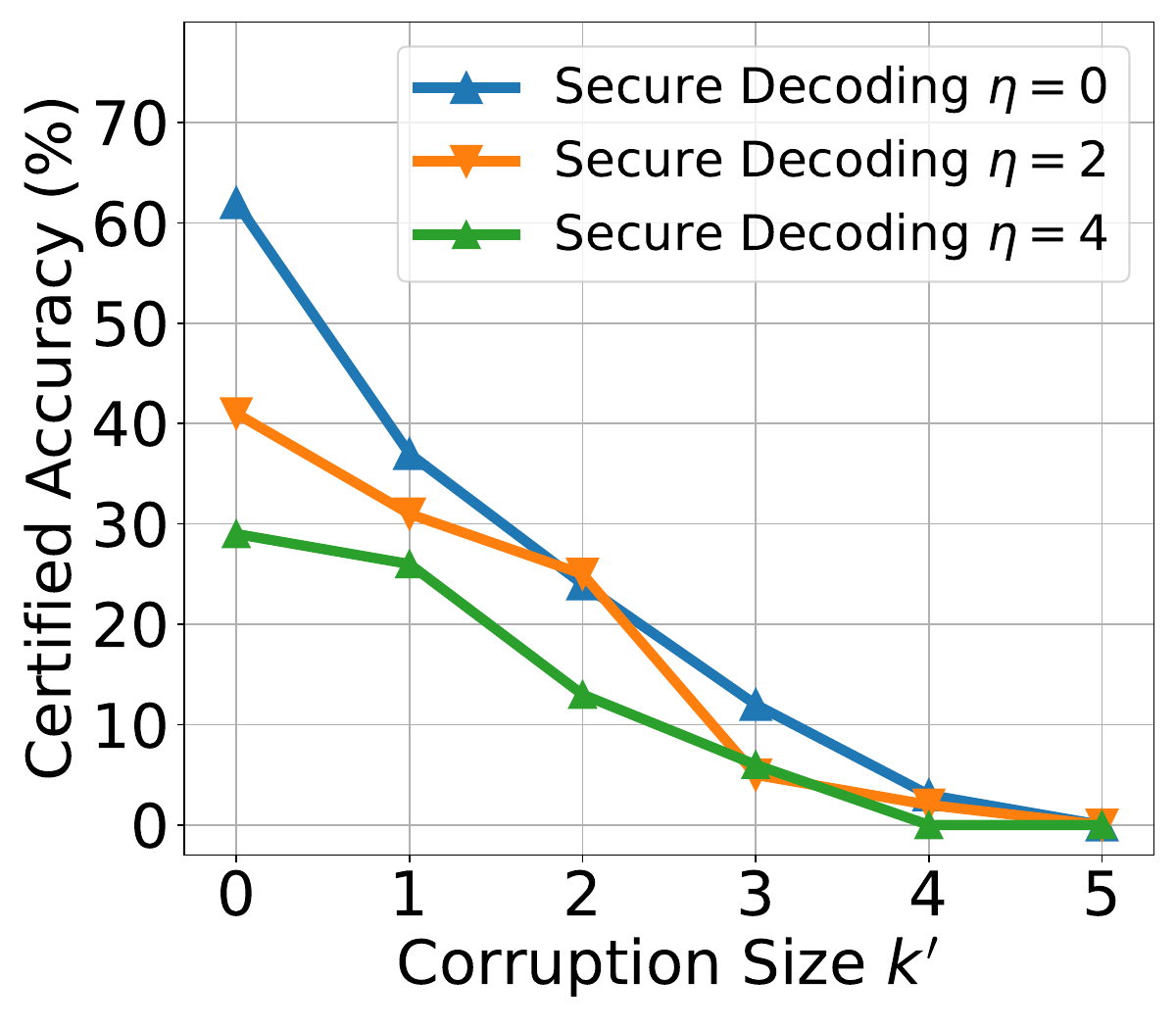}  \\
   (c) Natural Questions &  (d) RealtimeQA  \\
    \noalign{\medskip} \hdashline \noalign{\medskip}
   \includegraphics[width=0.4\linewidth]{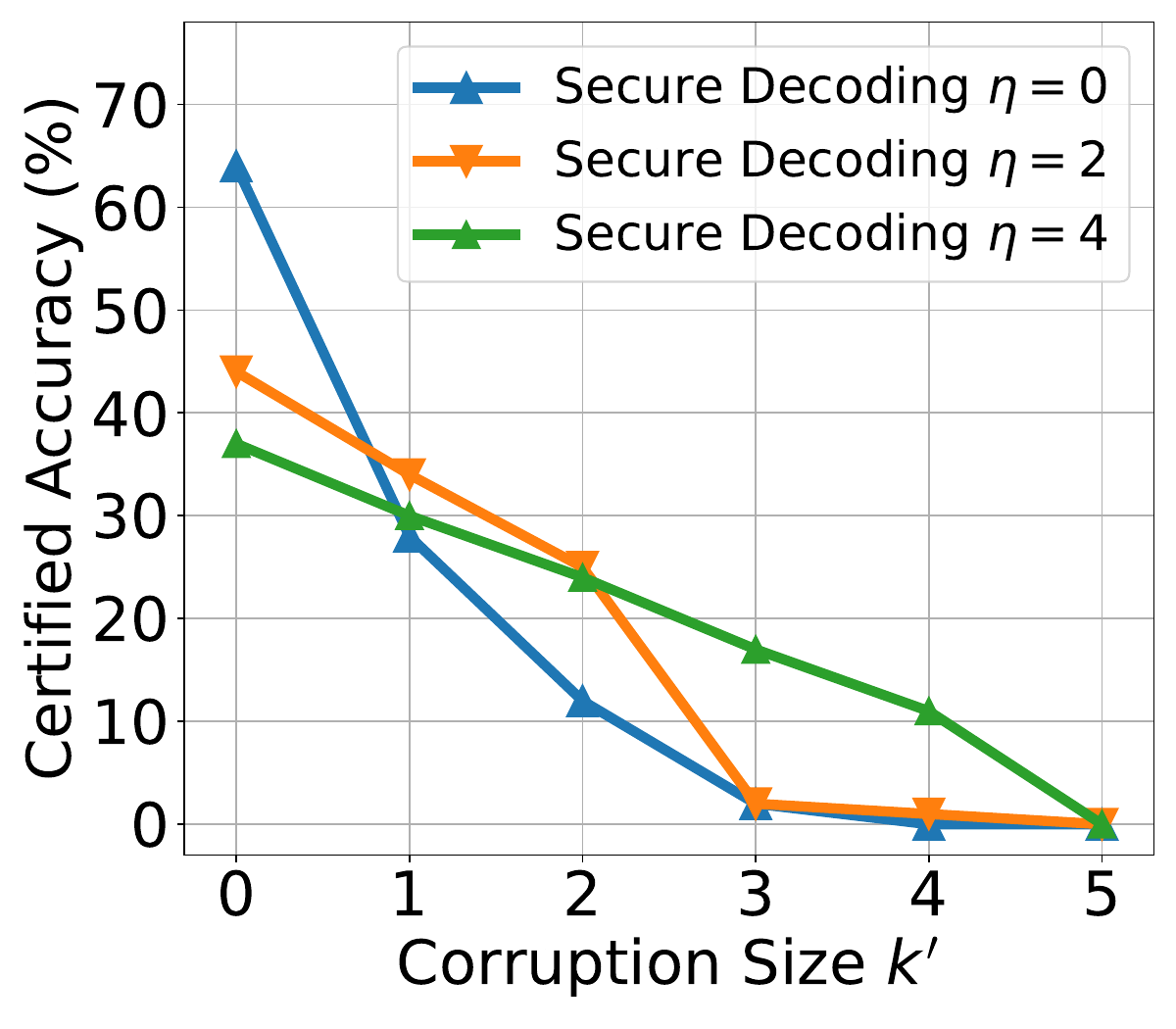} &
   \includegraphics[width=0.4\linewidth]{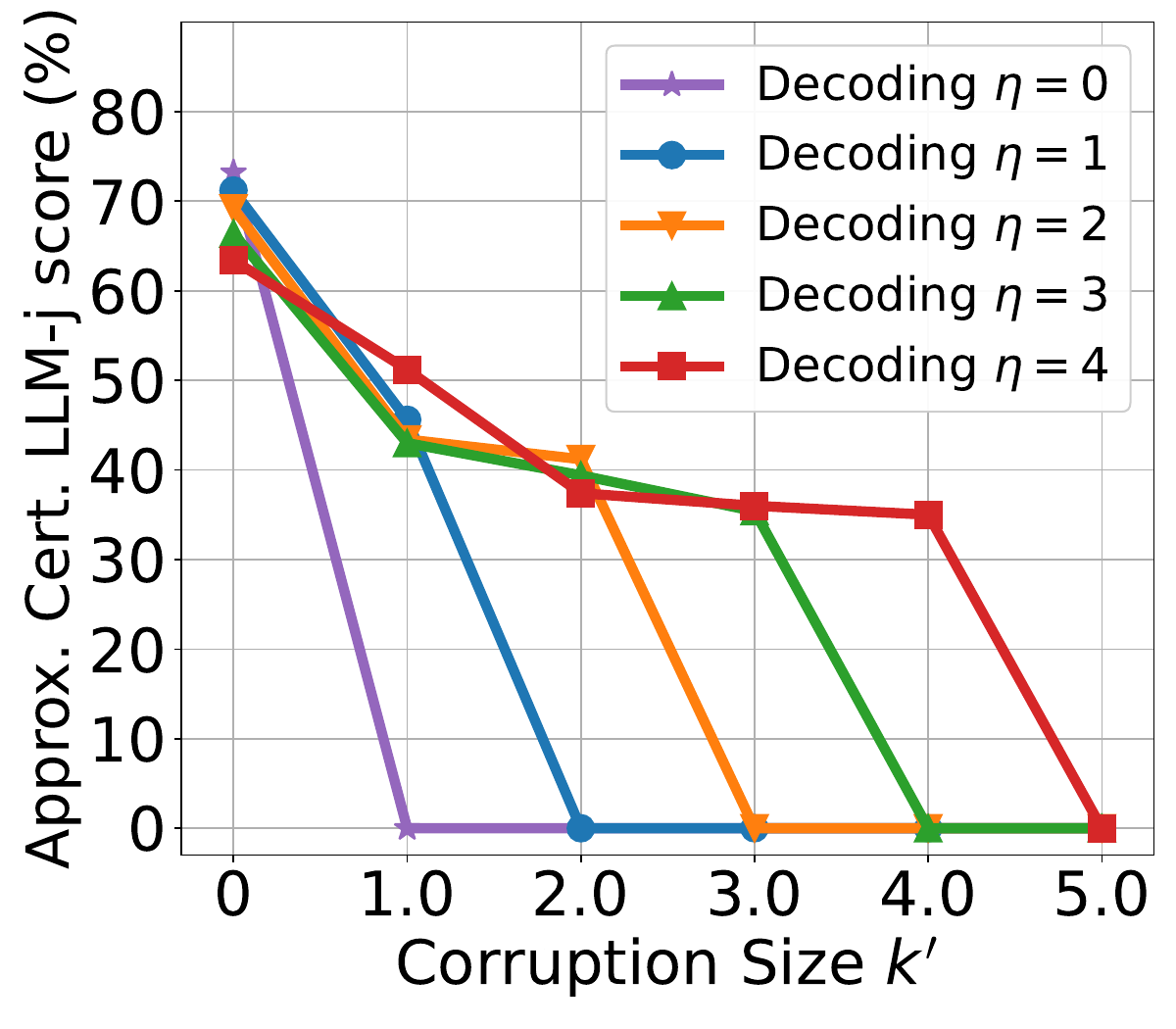}  \\
   (e) Natural Questions  &  (f) Biography generation 
 \end{tabular}}
 \caption{\framework robustness against different corruption sizes $k^\prime$ (Mistral-7B, $k=10$)}
 \label{Fig:apend_abl_corr}
\end{figure}

\textbf{Impact of corruption size $k^\prime$.} In Figure \ref{Fig:apend_abl_corr}, we report certifiable robustness for different corruption sizes $k^\prime$ using different \framework algorithms  and different datasets. \framework achieves substantial certifiable robustness even against multiple malicious passages. For instance, for the RealtimeQA-MC dataset (Figure~\ref{Fig:apend_abl_corr}(a)), the certifiable robust accuracy is still higher than 50\% when the corruption size is 3 out of 10. 

\textbf{Impact of keyword filtering thresholds $\alpha,\beta$.} In Figures \ref{Fig:apend_abl_corr}(b) and \ref{Fig:apend_abl_corr}(c), we report the robustness of keyword aggregation with different filtering thresholds $\alpha,\beta$. We can see that larger values of $\alpha,\beta$ are more robust to multiple-passages corruption, at the cost of a slight drop in benign performance (at corruption size $k^\prime=0$).

\textbf{Impact of decoding probability threshold $\eta$.} In Figures \ref{Fig:apend_abl_corr}(d) and \ref{Fig:apend_abl_corr}(e), we explore the effect of varying the decoding probability threshold $\eta$ on the RealtimeQA and Natural Questions datasets. We find that the benign accuracy (at $k^\prime=0$) drops as the $\eta$ increases; this is because a larger $\eta$ makes it more likely to output no-retrieval tokens and hurt performance. Interestingly, a larger $\eta$ can enhance robustness for Natural Questions in some cases (for larger corruption size $k^\prime$) but not for RealtimeQA. To explain this observation, we need to understand that, though a larger $\eta$ makes it more likely to form a finite response set $\cR$ during the certification (\textit{Case 4} is less likely to happen), the finite response set $\cR$ can contain responses made of more no-retrieval tokens, which might lead to lower $\tau$ values. Recall that Table~\ref{tab:main_cr} demonstrated that Mistral without retrieval performs much better on NQ (30\%) than RealtimeQA (8\%). This explains why Mistral can benefit more from a larger $\eta$ and more no-retrieval tokens on NQ, compared to RealtimeQA.  

In Figure~\ref{Fig:apend_abl_corr}(f), we further analyze $\eta$ for the biography generation task. As $\eta$ increases, the benign performance ($k^\prime=0$) decreases because \framework will output more non-retrieved tokens. However, a larger $\eta$ allows us to tolerate larger corruption size $k^\prime$, or $m^\prime$, because \textit{Case 4} (certification failure) will never happen when $\eta-m^\prime\geq0$ (recall Section~\ref{sec-cert-decoding}).



\section{Prompt Template} \label{apx-prompt}
We provide different prompt templates in this section.
\begin{figure}[H]
\setlength{\abovecaptionskip}{0pt}
\setlength\belowcaptionskip{0pt}
    \centering
\includegraphics[width=\linewidth]{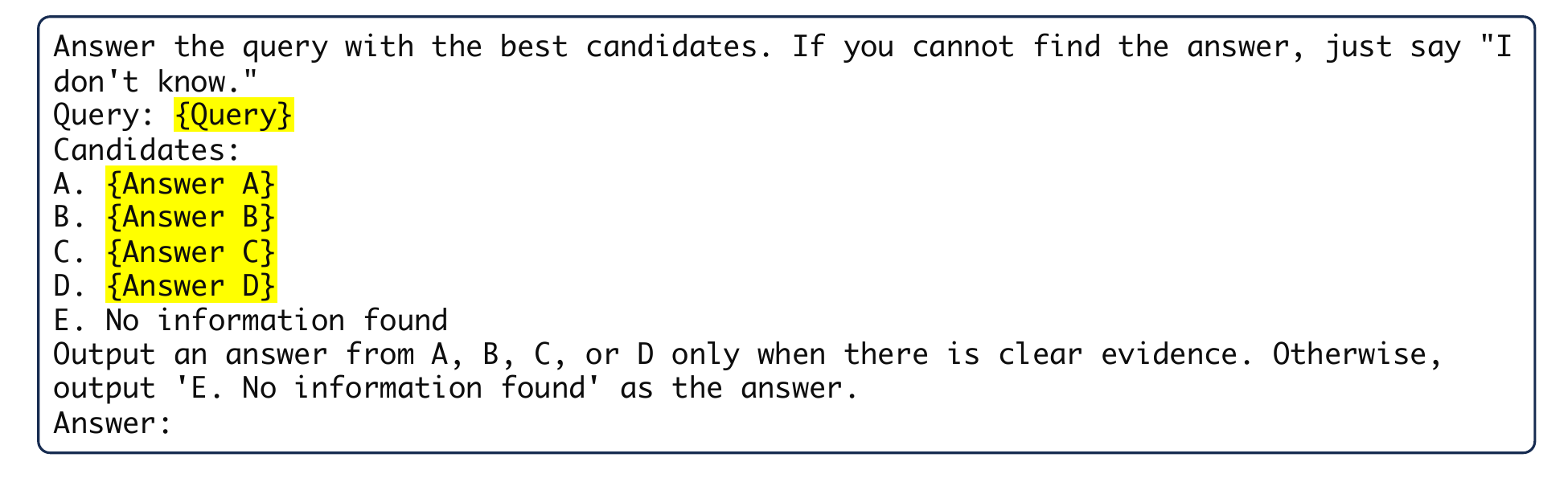}
    \caption{Template for multiple-choice QA without retrieval. }
    \label{fig-mc_nocontext}
\end{figure}

\begin{figure}[H]
\setlength{\abovecaptionskip}{0pt}
\setlength\belowcaptionskip{0pt}
    \centering
\includegraphics[width=\linewidth]{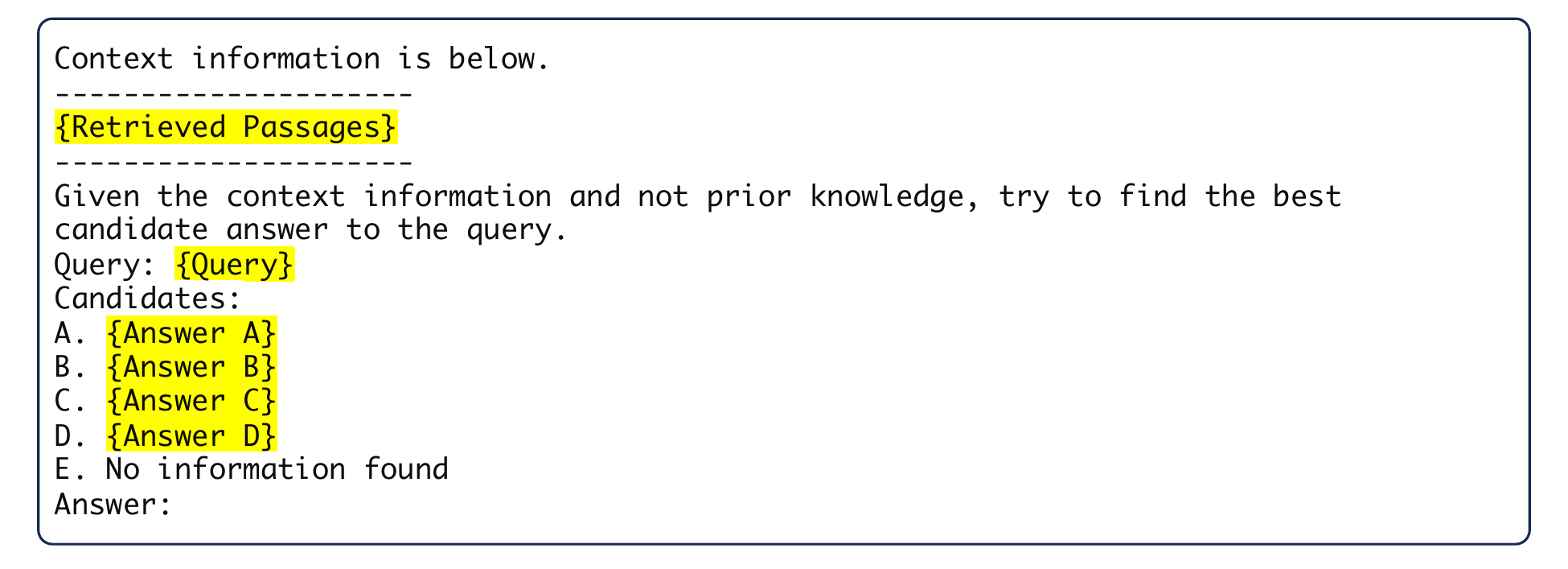}
    \caption{Template for multiple-choice QA with retrieval. }
    \label{fig-mc_context}
\end{figure}

\begin{figure}[H]
\setlength{\abovecaptionskip}{0pt}
\setlength\belowcaptionskip{0pt}
    \centering
\includegraphics[width=\linewidth]{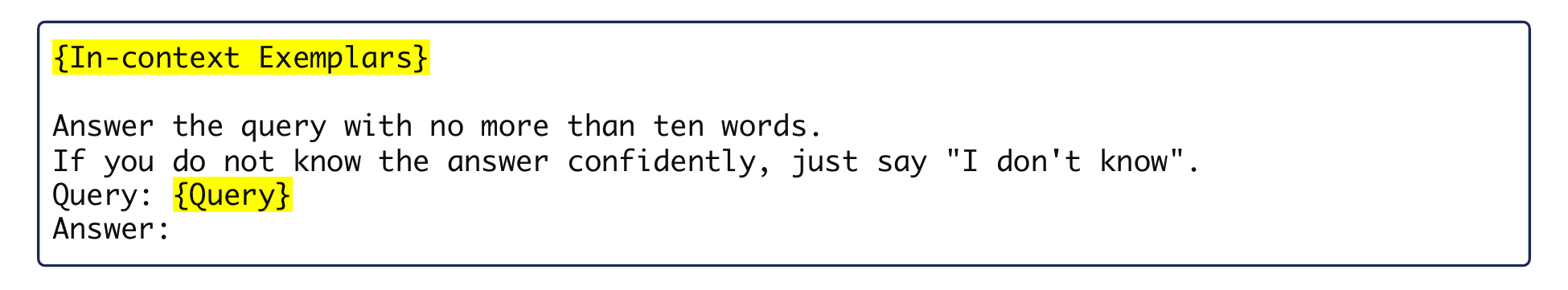}
    \caption{Template for open-domain QA without retrieval. }
    \label{fig-qa_nocontext}
\end{figure}

\begin{figure}[H]
\setlength{\abovecaptionskip}{0pt}
\setlength\belowcaptionskip{0pt}
    \centering
\includegraphics[width=\linewidth]{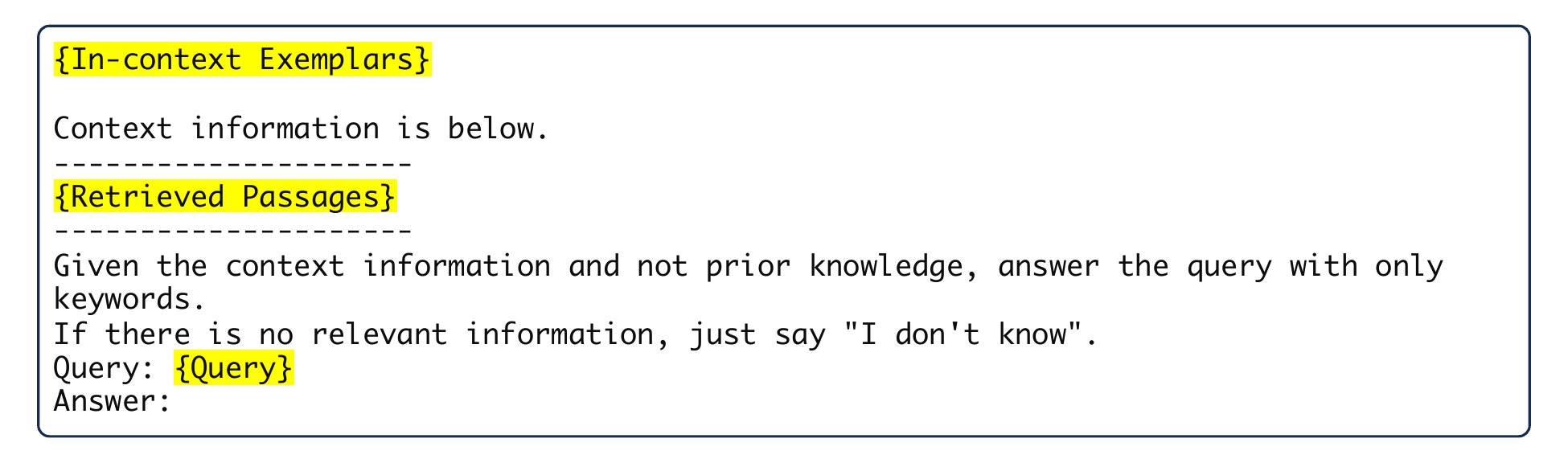}
    \caption{Template for open-domain QA with retrieval. }
    \label{fig-qa_context}
\end{figure}

\begin{figure}[H]
\setlength{\abovecaptionskip}{0pt}
\setlength\belowcaptionskip{0pt}
    \centering
\includegraphics[width=\linewidth]{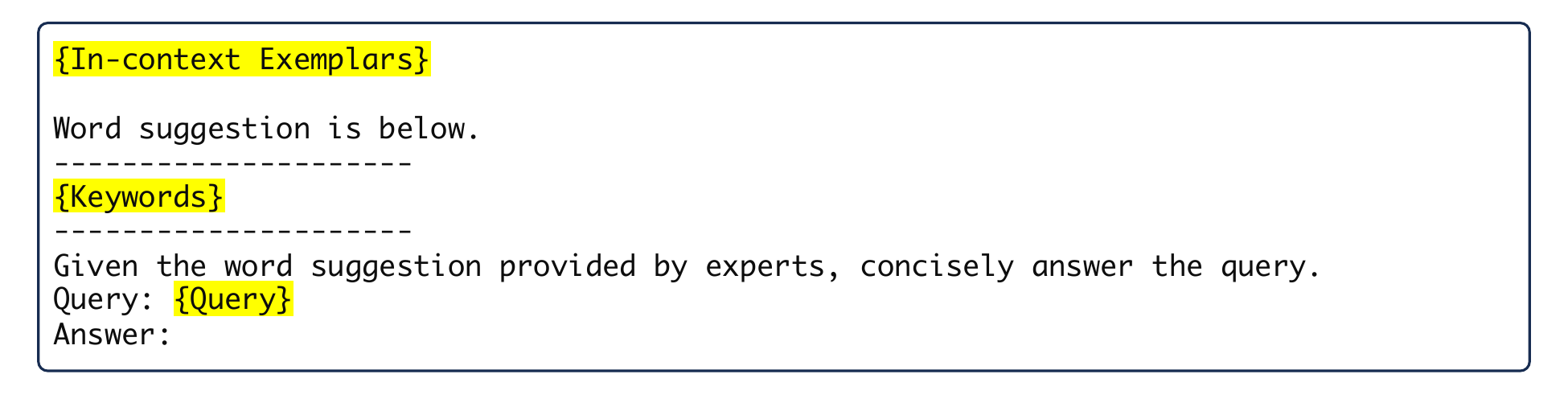}
    \caption{Template for keyword aggregation in open-domain QA. }
    \label{fig-qa_keyword}
\end{figure}

\begin{figure}[H]
\setlength{\abovecaptionskip}{0pt}
\setlength\belowcaptionskip{0pt}
    \centering
\includegraphics[width=\linewidth]{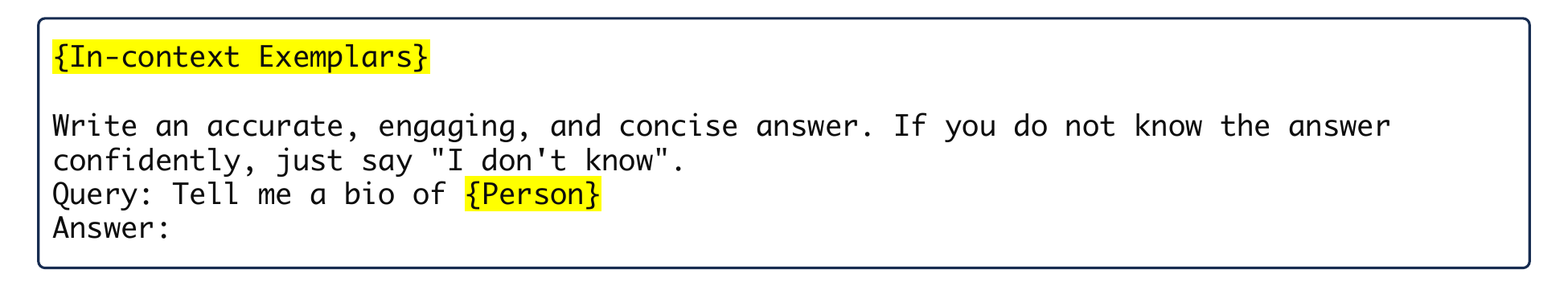}
    \caption{Template for biography generation without retrieval. }
    \label{fig-long_nocontext}
\end{figure}

\begin{figure}[H]
\setlength{\abovecaptionskip}{0pt}
\setlength\belowcaptionskip{0pt}
    \centering
\includegraphics[width=\linewidth]{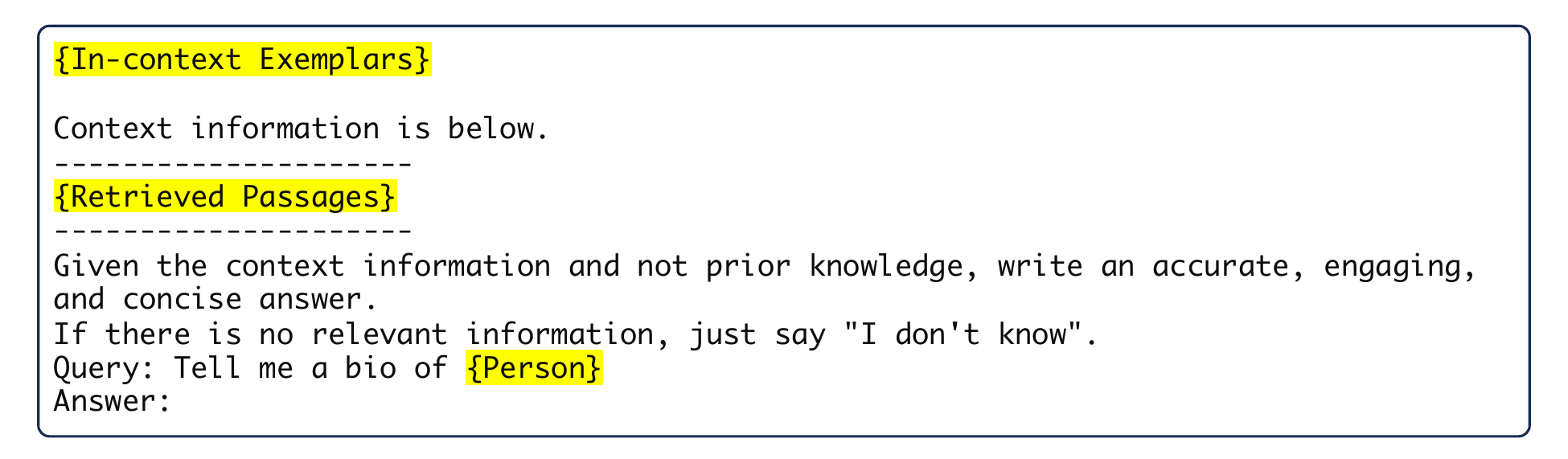}
    \caption{Template for biography generation with retrieval. }
    \label{fig-long_context}
\end{figure}

\begin{figure}[H]
\setlength{\abovecaptionskip}{0pt}
\setlength\belowcaptionskip{0pt}
    \centering
\includegraphics[width=\linewidth]{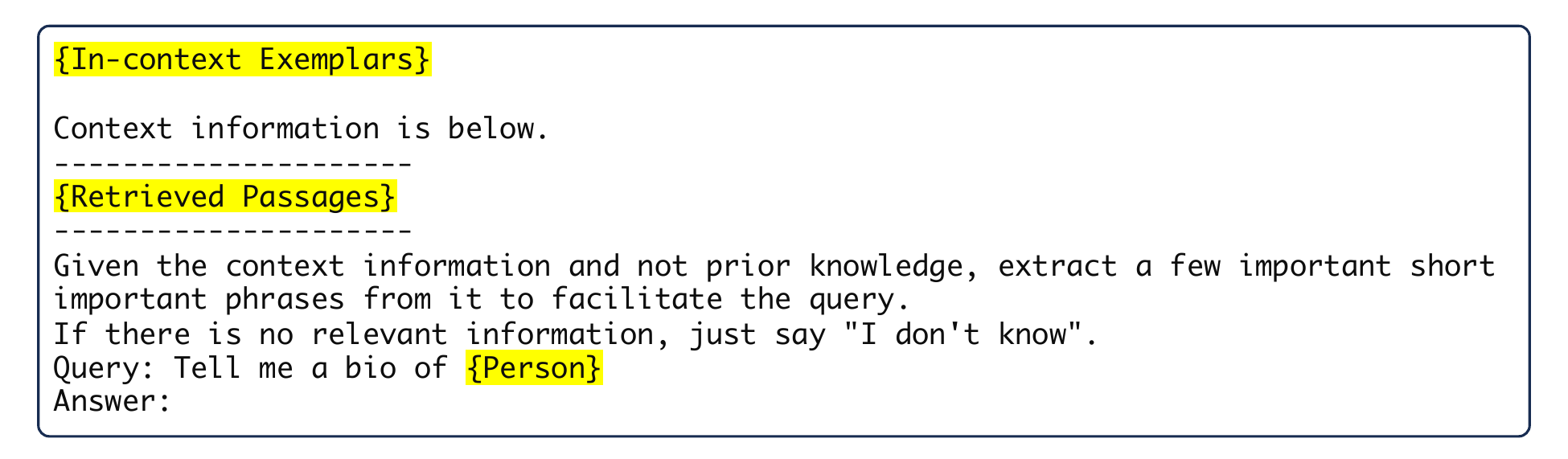}
    \caption{Template for generating keyword phases in biography generation. }
    \label{fig-long_genkeyword}
\end{figure}

\begin{figure}[H]
\setlength{\abovecaptionskip}{0pt}
\setlength\belowcaptionskip{0pt}
    \centering
\includegraphics[width=\linewidth]{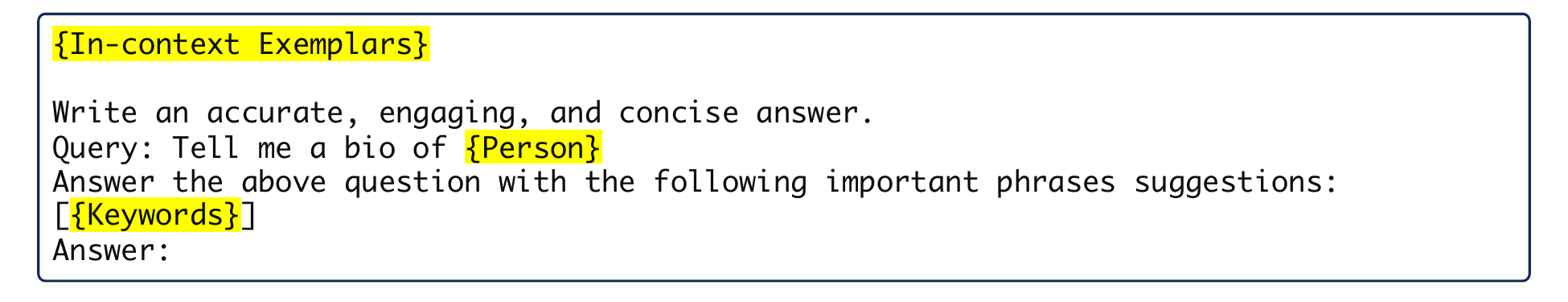}
    \caption{Template for keyword aggregation in biography generation. }
    \label{fig-long_keyword}
\end{figure}
\end{document}